\documentclass[10pt]{njulink}

\njulinksetup{
  type={Accepted by AAAI 2027},
  short-title={},
  subtitle={},
  keywords={}
}

\let\cite\citep
\usepackage{algorithm}
\usepackage{algorithmic}
\usepackage{amsmath}
\usepackage[table]{xcolor}

\title{AlphaCrafter: Harnessing Multi-Agent Workflows for Cross-Sectional Quantitative Trading}
\author[1]{Yishuo Yuan}
\author[1]{Jiayi Sheng}
\author[1]{Sirui Zeng}
\author[1]{Jiaqi Wang}
\author[1, \dag]{Jiaheng Liu}
\affiliation[1]{Nanjing University}
\contribution[\dag]{Corresponding author.}
\correspondence{Jiaheng Liu at \email{liujiaheng@nju.edu.cn}}
\date{October 2026}

\abstract{%
Quantitative trading agents have demonstrated substantial promise in automating factor discovery, signal aggregation, and portfolio execution. However, existing agent-based trading systems predominantly rely on loosely specified natural-language workflows, leading to opaque reasoning processes, inconsistent behaviors across foundation models, and limited controllability and verifiability, all of which introduce significant risks in financial decision-making. To address these limitations, we propose \textsc{AlphaCrafter}, a multi-agent framework built upon a structured agent harness. Instead of treating agent behavior as unconstrained prompt execution, AlphaCrafter encapsulates each agent within programmable policy specifications that integrate procedural workflows, execution constraints, and explicit verification mechanisms. This harness-driven design transforms the entire trading pipeline into a sequence of well-defined, reproducible, and auditable decision processes with explicit execution semantics. Extensive experiments on the CSI 300 and S\&P 500 benchmarks demonstrate that AlphaCrafter consistently achieves superior risk-adjusted returns while exhibiting substantially lower cross-model and cross-trial variance. These results suggest that harness-based agent design provides a practical foundation for building more reliable, controllable, and robust multi-agent systems for quantitative trading.
}

\begin{document}
\maketitle

\section{Introduction}
\label{sec:introduction}

Financial markets constitute high-dimensional, nonlinear dynamical systems characterized by heavy tails \cite{Mandelbrot1997}, volatility clustering \cite{6ab571e5-c8f0-3fcd-9005-ed6f2adc76d7}, and complex cross-sectional dependencies \cite{DIEBOLD2014119}. These properties imply that asset returns are jointly driven by macroeconomic regimes, microstructural frictions, and behavioral feedback \cite{038f59c7-dd05-312a-82e2-1c92036535ae, BROCK19981235, doi:10.1142/9789814417358_0006}. Consequently, the predictive power of any fixed signal set inevitably deteriorates as market conditions evolve. Nevertheless, the dominant paradigm in quantitative investing remains one of static model specification followed by periodic manual recalibration \cite{cao2025deeplearningllmssurvey, https://doi.org/10.1111/jofi.12080}.

Traditional quantitative strategies have evolved from classical factor pricing models \cite{https://doi.org/10.1111/j.1540-6261.1997.tb03808.x, FAMA19933} to machine learning methods such as XGBoost \cite{DBLP:journals/corr/ChenG16} and LightGBM \cite{10.5555/3294996.3295074}, and more recently to deep sequence architectures including LSTMs \cite{DBLP:journals/corr/abs-2105-06756} and Transformers \cite{DBLP:journals/corr/VaswaniSPUJGKP17}. Despite their impressive predictive capabilities, these approaches fundamentally operate under a static design philosophy: feature engineering, model architectures, and optimization objectives remain manually specified, requiring repeated human intervention whenever market dynamics shift.

The emergence of large language models (LLMs) has enabled agent-based systems capable of automating increasingly sophisticated components of quantitative trading. Existing approaches generally follow two directions. One line constructs role-playing trading committees that aggregate heterogeneous information through simulated discussions \cite{tian2025tradinggroupmultiagenttradingselfreflection, xiao2025tradingagentsmultiagentsllmfinancial}. While effective for financial reasoning, these anthropomorphic interactions introduce considerable latency and may amplify behavioral biases. Another line develops agentic frameworks that directly perform factor discovery, strategy optimization, or trading execution through iterative interactions with financial data and computational tools \cite{song2026trademinutesrationalitydrivenagentic, kou2025automatestrategyfindingllm, li2025rdagentquantmultiagentframeworkdatacentric, tang2025alphaagentllmdrivenalphamining}. Although these systems substantially automate quantitative research, they predominantly rely on loosely specified prompt-driven workflows that lack explicit execution policies, verification mechanisms, and behavioral consistency guarantees. These deficiencies undermine reproducibility and reliability---particularly in long-horizon trading, where sustained consistency is as critical as predictive accuracy.

Recent advances in agent engineering suggest that the execution harness surrounding an LLM can be as important as the underlying model itself for solving complex, long-running tasks \cite{openai2026harness, young2025effective}. Rather than relying solely on natural-language prompts, a well-designed harness specifies explicit execution policies and control mechanisms that transform agent behavior from an implicit prompt-driven process into a controllable, verifiable, and reproducible computational workflow.

Motivated by the principle of harness engineering, we propose \textsc{AlphaCrafter}, a harness-driven multi-agent framework for cross-sectional quantitative trading. AlphaCrafter encapsulates each agent within a structured execution harness that combines programmable policy specifications, execution constraints, and explicit verification mechanisms. The harness orchestrates the complete factor-to-execution pipeline while maintaining consistent behavior across different foundation models and repeated executions, enabling adaptive yet reliable quantitative trading under evolving market conditions.

Our core contributions are summarized as follows:

\begin{itemize}

\item \textbf{Harness-Driven Agent Execution for Quantitative Trading.}
We present the first harness-driven multi-agent framework for cross-sectional quantitative trading, replacing loosely specified prompt-driven workflows with programmable execution policies, explicit behavioral constraints, and verification mechanisms for controllable and reproducible agent execution.

% \item \textbf{Regime-Aware Dynamic Factor Construction.}
% We introduce a Screener harness that leverages the semantic reasoning capability of LLMs to identify prevailing market regimes and dynamically compose factor ensembles accordingly, enabling adaptive signal selection without retraining or manual intervention.

\item \textbf{Robust Long-Horizon Trading Performance.}
Extensive experiments on both CSI 300 and S\&P 500 demonstrate that AlphaCrafter consistently achieves superior risk-adjusted returns while exhibiting substantially lower cross-model and cross-trial variance, validating the effectiveness and robustness of harness-driven agent execution for quantitative trading.

\end{itemize}

\section{AlphaCrafter}
\label{sec:alphacrafter}

\subsection{Preliminary Formulation}
\label{subsec:preliminary}

\begin{figure}[t]
    \centering
    \includegraphics[width=\textwidth]{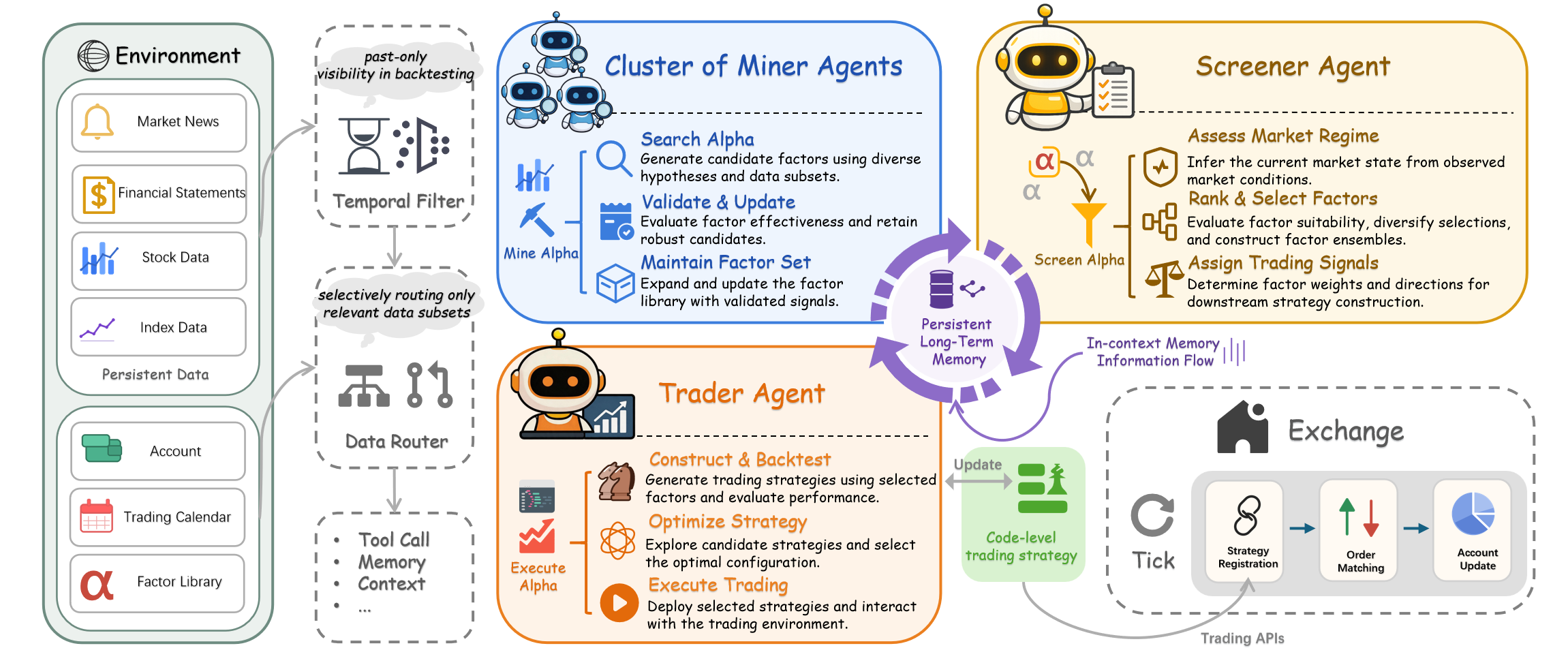}
    \caption{The architecture of AlphaCrafter: A Miner Cluster continuously expands the factor library, the Screener enforces regime-aware calibration, and the Trader adaptively optimizes execution strategy. Together, these three harnesses form a closed-loop trading system.}
    \label{fig:alphacrafter_architecture}
\end{figure}

The trading environment is formalized as a tuple
\begin{equation}
\mathcal{E} = (\mathcal{T}, \mathcal{M}, \mathcal{Z}, \Pi),
\end{equation}
where $\mathcal{T} = \{1, \ldots, T\}$ is the discrete set of trading days. $\mathcal{M}$ denotes the market state space, encompassing both aggregate market conditions and individual asset characteristics. Formally, $\mathcal{M}$ contains the universe of tradable assets $\mathcal{U} \subset \mathcal{M}$, where $\mathcal{U}$ is a set of $N$ individual entities. The raw market data is represented as a three-dimensional tensor $\mathbf{X} \in \mathbb{R}^{N \times T \times P}$, where $N$ denotes the number of assets, $T$ is the number of trading days, and $P$ denotes the number of raw features (e.g., open, high, low, close, volume). Each entry $x_{i,t}^{(p)}$ represents the value of the $p$-th feature for asset $i$ at time $t$. $\mathcal{Z}$ is the factor library, a dynamic repository of quantitative factors, where each factor is a function $f: \mathbb{R}^{\ell \times P} \to \mathbb{R}$ that maps a lookback window of raw features $\mathbf{x}_{i,t-\ell+1:t}$ for a single asset to a scalar signal, with $\ell$ denoting the lookback length. For a cross-section of $N$ assets, applying $f$ to each asset's historical feature window yields a cross-sectional signal vector $\mathbf{f}_t = (f(\mathbf{x}_{1,t-\ell+1:t}), \ldots, f(\mathbf{x}_{N,t-\ell+1:t})) \in \mathbb{R}^N$. We define an \textbf{alpha} as a factor that exhibits statistically significant predictive power for future cross-sectional returns, i.e., a factor whose information coefficient (IC) exceeds a predefined threshold. $\Pi$ defines the space of admissible trading strategies, parameterized by portfolio construction rules and risk constraints.

% $\mathcal{J}: \Pi \to \mathbb{R}$ is an evaluation functional that quantifies strategy performance, incorporating risk-adjusted return metrics.

Agents operate with two forms of memory. \textbf{In-context Memory} injects immediate situational awareness directly into each agent's LLM prompt. \textbf{Persistent Long-Term Memory} functions as a shared, universally accessible repository that archives the complete trading trajectory, including selected factors, executed strategies, and realised performance outcomes. This persistent layer enables collective system adaptation by disseminating performance feedback and conditional directives across all agents.

\subsection{Miner Cluster Harness}
\label{subsec:miner_cluster}
\small
\begin{algorithm}[t]
\caption{Miner Policy $P_M$}
\label{alg:miner_policy}
\begin{algorithmic}[1]
\STATE $\mathcal{Z}_{t+1} \gets \mathcal{Z}_t$
\STATE $\mathcal{H} \gets \{1, 5\}$ \COMMENT{Evaluation horizons}
\STATE $\Delta_{\text{valid}} \gets 90$ \COMMENT{Re-validation interval}

\STATE \textbf{FUNCTION} $\textsc{Validate}(f)$
    \FOR{each $h \in \mathcal{H}$}
        \STATE $\{\text{IC}_h^{(t)}\}_{t=1}^{T-h} \gets \emptyset$ \COMMENT{Initialize empty list of IC values}
        \FOR{$t = 1$ to $T-h$}
            \STATE $\mathbf{f}_t \gets (f(\mathbf{x}_{1,t-\ell+1:t}), \ldots, f(\mathbf{x}_{N,t-\ell+1:t})) \in \mathbb{R}^N$
            \STATE $R_{t+h} \gets (r_{1,t+h}, \ldots, r_{N,t+h}) \in \mathbb{R}^N$ \COMMENT{Forward returns: rate of price change from $t$ to $t+h$}
            \STATE $\text{IC}_h^{(t)} \gets \text{corr}(\mathbf{f}_t, R_{t+h})$
            \STATE $\text{RankIC}_h^{(t)} \gets \text{rank\_corr}(\mathbf{f}_t, R_{t+h})$
        \ENDFOR
        \STATE $\text{IC}_h \gets \text{mean}(\{\text{IC}_h^{(t)}\})$ \COMMENT{Time-series average of IC}
        \STATE $\text{RankIC}_h \gets \text{mean}(\{\text{RankIC}_h^{(t)}\})$ \COMMENT{Time-series average of Rank IC}
        \STATE $\text{ICIR}_h \gets \text{mean}(\{\text{IC}_h^{(t)}\}) / \text{std}(\{\text{IC}_h^{(t)}\})$ \COMMENT{ICIR}
        \STATE $\text{RankICIR}_h \gets \text{mean}(\{\text{RankIC}_h^{(t)}\}) / \text{std}(\{\text{RankIC}_h^{(t)}\})$ \COMMENT{Rank ICIR}
        \STATE $\eta_h \gets \frac{1}{T-h}\sum_{t=1}^{T-h} \mathbf{1}[\text{RankIC}_h^{(t)} > 0]$ \COMMENT{Hit ratio}
        \STATE $\kappa_h \gets \frac{1}{T-h}\sum_{t=1}^{T-h} \frac{|\{i \in [N] : f(\mathbf{x}_{i,t-\ell+1:t}) \neq \text{NA}\}|}{N}$ \COMMENT{Coverage}
        \STATE $\tau_h \gets \frac{1}{T-h-1}\sum_{t=2}^{T-h} \frac{\|\text{rank}(\mathbf{f}_t) - \text{rank}(\mathbf{f}_{t-1})\|_1}{N}$ \COMMENT{Turnover}
        \IF{$\big(\text{IC}_h, \text{RankIC}_h, \text{ICIR}_h, \text{RankICIR}_h, \eta_h, \kappa_h, \tau_h\big) \notin \mathcal{S}(h)$} 
            \STATE \textbf{return} False 
            \COMMENT{$\mathcal{S}(h)$: for $h=1$: $\text{IC},\text{RankIC}>0.015,\ \text{ICIR},\text{RankICIR}>0.2$; for $h=5$: $\text{IC},\text{RankIC}>0.025,\ \text{ICIR},\text{RankICIR}>0.25$; both: $|\eta-0.5|>0.1,\ \kappa>0.9,\ \tau<0.4$}
        \ENDIF
    \ENDFOR
    \STATE $\text{last\_validated}(f) \gets \text{current\_time}$
    \STATE \textbf{return} True
\STATE \textbf{END FUNCTION}

\STATE $\mathcal{C} \gets \textsc{LLMGenerateCandidates}(\mathcal{M})$ \COMMENT{LLM proposes prior hypotheses based on market conditions and generates multiple candidate factor expressions}

\FOR{each $f \in \mathcal{C}$}
    \IF{$\textsc{Validate}(f)$} \STATE $\mathcal{Z}_{t+1} \gets \mathcal{Z}_{t+1} \cup \{f\}$ \ENDIF
\ENDFOR

\FOR{each $f \in \mathcal{Z}_t$}
    \IF{$\text{current\_time} - \text{last\_validated}(f) \ge \Delta_{\text{valid}}$}
        \IF{$\textbf{not } \textsc{Validate}(f)$} \STATE $\mathcal{Z}_{t+1} \gets \mathcal{Z}_{t+1} \setminus \{f\}$ \ENDIF
    \ENDIF
\ENDFOR

\STATE \textbf{return} $\mathcal{Z}_{t+1}$
\end{algorithmic}
\end{algorithm}

The Miner Cluster harness is responsible for autonomous factor generation, rigorous validation, and continuous library maintenance. It deploys a cluster of specialized Miner agents that collaboratively expand the factor pool over the asset universe $\mathcal{U} \subset \mathcal{M}$. Candidate factors are proposed by an LLM based on current market conditions, evaluated against historical asset data, and selectively integrated into the factor library $\mathcal{Z}$. The harness terminates exploration after validating all generated factor candidates in the current round.

The Miner Cluster operates on the current factor library $\mathcal{Z}_t$, the asset universe $\mathcal{U}$, and market state $\mathcal{M}$:
\begin{equation}
A_M(\mathcal{Z}_t, \mathcal{M}) \xrightarrow{P_M} \mathcal{Z}_{t+1}.
\end{equation}

The policy $P_M$ is summarized in Algorithm~\ref{alg:miner_policy}. It begins by invoking the agent to generate candidate factor expressions conditioned on the prevailing market state. Each candidate is then validated using a comprehensive set of metrics including the Information IC, Rank IC, ICIR, Rank ICIR, hit ratio, coverage, and turnover, evaluated across multiple prediction horizons. Accepted factors are persisted into the library, while a subsequent maintenance phase re-validates existing factors at regular intervals and prunes those exhibiting significant performance decay.

\subsection{Screener Harness}
\label{subsec:screener_agent}

The Screener harness distills a coherent factor ensemble $\mathcal{E}_t$ by weaving together signals from the factor library with a nuanced reading of prevailing market conditions. Its policy $P_S$ first retrieves market information via a dedicated tool, then invokes an LLM to diagnose the market regime $\hat{\mathcal{R}}_t$, assessing trend direction, volatility, correlation structure, and other relevant market characteristics. Building on this regime diagnosis, it evaluates the semantic relevance of each factor $f \in \mathcal{Z}_t$ with respect to the diagnosed regime, computes a suitability score based on recent predictive performance, and selectively assembles a diversified subset by enforcing semantic dissimilarity among selected factors. Finally, it assigns directional weights to form an ensemble attuned to the unfolding market dynamics.

The Screener's transformation of the factor library and market state into an actionable ensemble is given by:
\begin{equation}
A_S(\mathcal{Z}_t, \mathcal{M}) \xrightarrow{P_S} (\mathcal{E}_t, \hat{\mathcal{R}}_t).
\end{equation}

Algorithm~\ref{alg:screener_policy} outlines the selection process. The agent retrieves market information, diagnoses the regime via LLM, filters factors by semantic relevance, computes suitability scores as the mean of recent Rank IC values, enforces diversity through LLM-based semantic similarity assessment, and outputs a structured ensemble with normalized weights and directional signs.

\begin{algorithm}[t]
\small
\caption{Screener Policy $P_S$}
\label{alg:screener_policy}
\begin{algorithmic}[1]
\STATE $\text{Info}_t \gets \textsc{CallGetMarketInfoTool}(\mathcal{M})$ \COMMENT{Retrieve market data and information through dedicated data acquisition tools}
\STATE $\hat{\mathcal{R}}_t \gets \textsc{LLMAssessRegime}(\text{Info}_t)$ \COMMENT{LLM diagnoses market regime based on retrieved information}
\STATE $\text{selected} \gets \emptyset,\;\; \text{candidate} \gets \emptyset$

\FOR{each $f \in \mathcal{Z}_t$}
    \STATE $\text{relevant} \gets \textsc{LLMSemanticJudge}(f, \hat{\mathcal{R}}_t)$ \COMMENT{LLM judges semantic relevance; skip if irrelevant}
    \IF{$\text{relevant} = \text{False}$} \STATE \textbf{continue} \ENDIF
    \STATE $\mathbf{f}_\tau \gets (f(\mathbf{x}_{1,\tau-\ell+1:\tau}), \ldots, f(\mathbf{x}_{N,\tau-\ell+1:\tau})) \in \mathbb{R}^N$ \COMMENT{Factor vector over cross-section}
    \STATE $s_f \gets \frac{1}{L_{\text{ic}}}\sum_{\tau=t-L_{\text{ic}}}^{t-1} \text{rank\_corr}(\mathbf{f}_\tau, R_{\tau+1})$ \COMMENT{Mean Rank IC over the past $L_{\text{ic}}$ days as factor suitability score for current market conditions}
    \STATE $\text{candidate} \gets \text{candidate} \cup \{(f, s_f)\}$
\ENDFOR

\STATE $\text{ranked} \gets \text{sort}(\text{candidate}, |s_f|, \text{descending})$

\WHILE{$\text{ranked} \neq \emptyset$}
    \STATE $f^* \gets \text{pop\_first}(\text{ranked})$
    \IF{$|s_{f^*}| < \theta_s$} \STATE \textbf{break} \ENDIF \COMMENT{Reject factors with insufficient predictive strength; $\theta_{s}=0.02$}
    \STATE $\text{similarity} \gets \max_{g \in \text{selected}} \textsc{LLMSemanticSimilarity}(f^*, g)$ \COMMENT{LLM evaluates semantic similarity between factor expressions following the given instruction; returns 0 if selected is empty}
    \IF{$\text{similarity} < \theta_{\text{sim}}$} 
    \STATE $\text{selected} \gets \text{selected} \cup \{f^*\}$
    \COMMENT{$\theta_{\text{sim}} = 0.8$: LLM-guided semantic filtering, reject if factors are semantically overlapping}
    \ENDIF
\ENDWHILE

\FOR{each $f \in \text{selected}$}
    \STATE $w_f \gets \frac{|s_f|}{\sum_{g \in \text{selected}} |s_g|}$ \COMMENT{Weight normalized by absolute suitability}
    \STATE $d_f \gets \text{sign}(s_f)$ \COMMENT{Long if $s_f>0$, short if $s_f<0$}
    \STATE $\mathcal{E}_t \gets \mathcal{E}_t \cup \{(f, w_f, d_f)\}$
\ENDFOR

\STATE \textbf{return} $(\mathcal{E}_t, \hat{\mathcal{R}}_t)$
\end{algorithmic}
\end{algorithm}

\subsection{Trader Harness}
\label{subsec:trader_agent}

The Trader harness composes a trading strategy $\pi_t$ by integrating the factor ensemble $\mathcal{E}_t$ and regime assessment $\hat{\mathcal{R}}_t$ with self-determined hyperparameter configurations, operating on the asset universe $\mathcal{U}$. Its policy $P_T$ explores hyperparameter configurations $\Theta$ of a reference top-$K$ strategy $\pi_{\text{ref}}$ via backtesting, selects the configuration that satisfies the feasible region $\mathcal{S}_{\Theta}$ while maximizing the Sharpe ratio, and executes the resulting portfolio.

The Trader's operation is formalized as:
\begin{equation}
A_T(\mathcal{E}_t, \hat{\mathcal{R}}_t, \mathcal{M}) \xrightarrow{P_T} (\pi_t, r_t),
\end{equation}
where $r_t$ is the realized return from executing $\pi_t$.

The reference strategy $\pi_{\text{ref}}$ is a top-$K$ long-short portfolio construction mechanism operating on the asset universe $\mathcal{U}$, detailed in Algorithm~4 of Appendix~\ref{sec:reference_strategy}. For each asset, it aggregates weighted signals from the factor ensemble $\mathcal{E}_t$ into a composite score, then selects the top $N_{\text{long}}$ assets for long positions and the bottom $N_{\text{short}}$ for short positions. Position sizing follows a controlled allocation scheme parameterized by gross exposure $\beta$ and net exposure bias $\gamma$, with rebalancing subject to these exposure constraints.

Algorithm~\ref{alg:trader_policy} describes the strategy search and execution policy. The agent samples hyperparameter configurations conditioned on the diagnosed market regime, generates executable strategy code by instantiating the reference strategy with the factor ensemble and sampled parameters, evaluates candidates via backtesting, and selects the configuration that achieves the highest Sharpe ratio while satisfying risk constraints. The optimal strategy is then executed on live market data to produce realized returns.

\begin{algorithm}[t]
\small
\caption{Trader Policy $P_T$}
\label{alg:trader_policy}
\begin{algorithmic}[1]
\STATE $\Theta \gets \{\beta, \gamma, N_{\text{long}}, N_{\text{short}}\}$ \COMMENT{Strategy hyperparameters}
\STATE $N \gets 3$ \COMMENT{Maximum backtest trials}
\STATE $\pi_{\text{best}} \gets \text{None},\;\; \text{SR}_{\text{best}} \gets -\infty$

\FOR{$n \gets 1$ \TO $N$}
    \STATE $\theta_n \gets \textsc{Sample}(\Theta, \hat{\mathcal{R}}_t)$ \COMMENT{Sample hyperparameters conditioned on market regime to adjust long/short intensity and position sizing}
    \STATE $\pi_n \gets \textsc{LLMGenerateCode}(\pi_{\text{ref}}, \mathcal{E}_t, \theta_n)$ \COMMENT{Generate executable strategy code}
    \STATE $r_n, \text{SR}_n, \text{MDD}_n \gets \textsc{CallBacktestTool}(\pi_n)$
    \IF{$\big(r_n, \text{SR}_n, \text{MDD}_n\big) \in \mathcal{S}$ \AND $\text{SR}_n > \text{SR}_{\text{best}}$}
    \STATE $\pi_{\text{best}} \gets \pi_n,\;\; \text{SR}_{\text{best}} \gets \text{SR}_n$
    \COMMENT{$\mathcal{S}$: $r > 8\%,\ \text{SR} > 0.6,\ \text{MDD} > -8\%$}
    \ENDIF
\ENDFOR

\IF{$\pi_{\text{best}} = \text{None}$} \STATE \textbf{return} $(\text{None}, 0)$ \ENDIF

\STATE $r_t \gets \textsc{CallExecuteTool}(\pi_{\text{best}})$ \COMMENT{Run the generated strategy code on live market data}
\STATE \textbf{return} $(\pi_{\text{best}}, r_t)$
\end{algorithmic}
\end{algorithm}

%---------------------------

% \subsection{Overall System Objective}
% \label{subsec:objective}
% Formally, at each decision trading day $t$, the composite action of the harnesses yields a strategy $\pi_t \in \Pi$. The objective is to maximize the evaluation functional $\mathcal{J}$ applied to the strategy executed in the subsequent period:
% \begin{equation}
% \max_{\pi_t \in \Pi_t} \mathbb{E} \left[ \mathcal{J}(\pi_t) \right],
% \end{equation}
% where $\Pi_t \subseteq \Pi$ is the subset of admissible strategies realizable given the current factor library $\mathcal{Z}_t$ and market regime $\hat{\mathcal{R}}_t$. The functional $\mathcal{J}$ inherently penalizes drawdowns and volatility, aligning with the goal of stable capital appreciation over the trading horizon $T$.

\section{Experiments}
\label{sec:experiments}

\subsection{Experimental Setup}
\label{subsec:experimental_setup}

\subsubsection{Dataset}
% cite: CSI 300 index
% cite: S&P 500 index
% cite: price-volume, fundamental, financial statement, alternative data sources
Our experiments utilize a comprehensive dataset covering both Chinese A-share market (CSI 300 constituents) and U.S. stock market (S\&P 500 constituents). The raw data encompasses four categories: (1) \textbf{Price-volume data} including daily OHLCV (Open, High, Low, Close, Volume); (2) \textbf{Fundamental indicators} including Price-to-Earnings (PE), Price-to-Sales (PS), Price-to-Book (PB), and Dividend Yield Rate (DYR); (3) \textbf{Financial statements} including quarterly balance sheets, income statements, and cash flow statements; (4) \textbf{Alternative data} comprising financial news and corporate announcements, including authoritative sources such as the Federal Reserve. The temporal split of the dataset is detailed in Table~\ref{tab:data_split}. Detailed information regarding data sources and storage formats is provided in Appendix~\ref{subsec:dataset_details}.

\begin{table}[h]
\small
\centering
\caption{Dataset Splits for Training, Validation, Backtesting, and Live Trading}
\label{tab:data_split}
\renewcommand{\arraystretch}{1.05}
\setlength{\tabcolsep}{4pt}
\begin{tabular}{l|l|l|l|l}
\hline
\textbf{Market} & \textbf{Training} & \textbf{Validation} & \textbf{Backtesting} & \textbf{Live Trading} \\
\hline
\textbf{CSI 300} & 2016.01.04--2022.12.30 & 2023.01.03--2023.12.29 & 2024.01.02--2026.02.27 & 2026.03.02--2026.06.12 \\
\textbf{Trading Days} & 1703 Days & 242 Days & 519 Days & 69 Days \\
\hline
\textbf{S\&P 500} & 2016.01.04--2022.12.30 & 2023.01.03--2023.12.29 & 2024.01.02--2026.02.27 & 2026.03.02--2026.06.12 \\
\textbf{Trading Days} & 1763 Days & 250 Days & 542 Days & 73 Days \\
\hline
\end{tabular}
\end{table}

\subsubsection{Metrics}
\label{subsubsec:metrics}

We evaluate all methods using three standard financial metrics. \textbf{Annualized Return (AR)} measures the geometric mean yearly return, reflecting absolute profitability. \textbf{Sharpe Ratio (SR)} assesses risk-adjusted performance by normalizing excess returns by their volatility, capturing return efficiency per unit of risk. \textbf{Maximum Drawdown (MDD)} quantifies the largest peak-to-trough decline over the evaluation period, indicating worst-case capital loss. For factor-level evaluation, we report \textbf{Information Coefficient (IC)} and \textbf{IC Information Ratio (ICIR)}, along with their rank-based variants. IC measures the cross-sectional Pearson correlation between factor values and forward returns, directly indicating predictive strength. Rank IC uses Spearman rank correlation instead, providing robustness against outliers. ICIR and Rank ICIR are computed as the respective mean IC divided by its standard deviation, measuring the consistency of predictive performance over time. Detailed formal definitions of all metrics are provided in Appendix~\ref{subsec:metrics_details}.

\subsubsection{Baselines}
\label{subsubsec:baselines}

We compare AlphaCrafter against representative methods spanning five categories. \textbf{Quantitative methods} include MACD \cite{appel1979macd}, a momentum-based trend-following indicator, and Grid Trading \cite{0f43c3bb-1dea-3498-87be-cd7d263d30dc}, a mean-reversion strategy that places buy and sell orders at predetermined price levels. \textbf{Machine learning methods} employ LightGBM \cite{10.5555/3294996.3295074} and XGBoost \cite{DBLP:journals/corr/ChenG16} trained on technical and fundamental features to predict asset returns. \textbf{Deep learning methods} include LSTM \cite{DBLP:journals/corr/abs-2105-06756} and Transformer \cite{DBLP:journals/corr/VaswaniSPUJGKP17}, which capture temporal dependencies through sequential modeling, as well as TRA \cite{lin2021tra}, which additionally captures cross-sectional patterns via a dual-attention architecture. \textbf{Traditional trading agent methods} comprise TradingAgents \cite{xiao2025tradingagentsmultiagentsllmfinancial} and TradingGroup \cite{tian2025tradinggroupmultiagenttradingselfreflection}, which simulate role-playing committees with rule-based coordination for multi-source information aggregation. \textbf{Quantitative trading agent methods} include RD-Agent \cite{li2025rdagentquantmultiagentframeworkdatacentric} and AlphaAgent \cite{tang2025alphaagentllmdrivenalphamining}, which leverage LLMs for automated factor generation. Beyond the multi-agent collaborative aggregation of market information, AlphaAgent further incorporates a regularization-based search mechanism to counteract alpha decay. All baseline implementations follow their original papers' recommended settings, with hyperparameters tuned on the validation set. For methods requiring feature preprocessing, input data are standardized using Z-score normalization. For LLM-based agent methods, we conduct experiments using three backbone models: \textbf{GPT 5.3 Codex} \cite{openai2026gpt53codex}, \textbf{Claude Opus 4.6} \cite{anthropic2026claudeopus46}, and \textbf{Gemini 3.1 Pro} \cite{googledeepmind2026gemini31pro}.

\subsubsection{Settings}
\label{subsubsec:settings}
Our experiments are conducted under a daily-frequency cross-sectional trading framework with portfolio weights updated at each market close. To mitigate the risk of data leakage, all stock identifiers are anonymized throughout the experiments, so that LLM-based methods cannot exploit memorized firm-specific information from pre-training. For LLM-based methods, each configuration is evaluated over 10 independent trials; reported metrics are averaged over trials within the interquartile range to mitigate outlier influence. Backtesting is performed through a simulation environment with static market data and a simulated matching engine calibrated to real liquidity conditions. We additionally conduct a live-trading phase using a paper-trading API from a real brokerage, operating under actual market order execution mechanics. The evaluation window for live trading falls strictly outside the training data cutoff of all backbone models to eliminate confounding effects from LLM memory and known market-beta trends \cite{kong2026evaluatingllmsfinancerequires}. Detailed experimental specifications are deferred to Appendix~\ref{subsec:simulation_settings}.

\subsection{Main Results}
\label{subsec:main_result}

Table~\ref{tab:combined} reports the backtesting and live trading performance of all evaluated methods across the CSI~300 and S\&P~500 markets.

\newcommand{\best}[1]{\textcolor[HTML]{1A7A4C}{\textbf{#1}}}
\newcommand{\good}[1]{\textcolor[HTML]{95D5B2}{\textbf{#1}}}

\begin{table}[htbp]
\centering
\caption{Backtesting and Live Trading Performance Comparison on CSI 300 and S\&P 500}
\label{tab:combined}
\scriptsize
\setlength{\tabcolsep}{4pt}
\renewcommand{\arraystretch}{1.08}
\begin{tabular*}{\textwidth}{l|ccc|ccc|ccc|ccc}
\hline
 \multirow{3}{*}{\textbf{Method}} 
& \multicolumn{6}{c|}{\textbf{Backtesting}} 
& \multicolumn{6}{c}{\textbf{Live Trading}} \\
\cline{2-13}
& \multicolumn{3}{c|}{\textbf{CSI 300}} & \multicolumn{3}{c|}{\textbf{S\&P 500}}
& \multicolumn{3}{c|}{\textbf{CSI 300}} & \multicolumn{3}{c}{\textbf{S\&P 500}} \\
\cline{2-13}
& \textbf{AR(\%)} & \textbf{SR} & \textbf{MDD(\%)} & \textbf{AR(\%)} & \textbf{SR} & \textbf{MDD(\%)}
& \textbf{AR(\%)} & \textbf{SR} & \textbf{MDD(\%)} & \textbf{AR(\%)} & \textbf{SR} & \textbf{MDD(\%)} \\
\hline
\rowcolor[HTML]{F5E4D6} \multicolumn{13}{l}{\textbf{Non-LLM Methods}} \\
\hline
MACD & \best{21.77} & 0.9856 & -18.63 & 7.92 & 0.7203 & -19.88 & 10.29 & 0.8662 & -16.18 & \best{18.76} & 1.2209 & -19.47 \\
Grid Trading & 6.62 & 0.3291 & -16.02 & 15.21 & 1.1115 & -9.18 & -2.90 & -0.2552 & -12.92 & 6.22 & 0.3901 & -14.70 \\
LightGBM & 17.92 & 1.2215 & -10.32 & 8.14 & 0.5142 & -13.67 & 7.64 & 0.5140 & -9.70 & 11.99 & 1.1892 & -12.15 \\
XGBoost & 16.26 & 1.2431 & \good{-8.71} & 2.08 & -0.1625 & -8.63 & 3.40 & 0.3621 & -10.17 & 9.68 & 1.0110 & -13.32 \\
LSTM & \good{20.11} & 1.4279 & -12.21 & \best{16.26} & \best{1.3802} & -10.46 & 3.22 & 0.3469 & -11.59 & 7.52 & 0.8262 & -10.91 \\
Transformer & 14.31 & 1.1917 & -9.17 & 7.22 & 0.4420 & -10.11 & -2.31 & -0.3223 & -13.92 & 5.09 & 0.2826 & -13.15 \\
TRA & 16.25 & 1.1622 & -10.65 & 10.27 & 1.0071 & -8.19 & 7.51 & 0.7121 & -8.87 & 9.07 & 1.0225 & -10.05 \\
\hline
\rowcolor[HTML]{E8F0E0} \multicolumn{13}{l}{\textbf{Agent Methods with GPT 5.3 Codex}} \\
\hline
TradingAgents & 15.25 & 1.2726 & -9.05 & 10.75 & 0.9769 & -11.40 & 7.72 & 0.5673 & -12.84 & 10.45 & 1.1243 & -11.61 \\
TradingGroup & 13.43 & 1.1631 & -10.04 & 11.67 & 1.0431 & -9.83 & 4.66 & 0.3005 & -12.68 & 14.20 & 1.2978 & -13.02 \\
RD-Agent & 16.32 & 1.3935 & -9.04 & 9.26 & 0.8512 & \best{-7.95} & 5.98 & 0.3571 & -11.92 & 9.21 & 1.1871 & -11.84 \\
AlphaAgent & 16.11 & 1.5043 & -9.55 & 15.22 & 1.3123 & -9.22 & \good{10.78} & 1.1035 & -10.66 & 12.23 & 1.1341 & -12.35 \\
AlphaCrafter & 16.76 & 1.5268 & -8.98 & 13.51 & 1.2531 & -8.85 & 9.57 & 1.1275 & -9.07 & 14.02 & 1.4546 & \good{-9.06} \\
\hline
\rowcolor[HTML]{D6E6DF} \multicolumn{13}{l}{\textbf{Agent Methods with  Claude Opus 4.6}} \\
\hline
TradingAgents & 17.25 & 1.3726 & -9.78 & 11.21 & 1.0769 & -10.08 & 6.72 & 0.6671 & -10.22 & 13.45 & 1.2521 & -13.10 \\
TradingGroup & 13.53 & 1.1368 & -12.65 & 10.58 & 0.9813 & -11.26 & 3.52 & 0.4998 & -8.05 & 8.32 & 0.9186 & -10.31 \\
RD-Agent & 16.48 & 1.3235 & -10.14 & 11.48 & 1.1334 & -9.34 & 5.52 & 0.5164 & -10.53 & 8.12 & 0.7810 & -11.96 \\
AlphaAgent & 19.22 & \good{1.6648} & -8.92 & 14.51 & 1.2664 & -9.17 & \best{11.22} & 1.1783 & -9.92 & 15.75 & \good{1.5767} & -9.76 \\
\rowcolor[HTML]{E9F7EA} AlphaCrafter & 18.88 & \best{1.6732} & \best{-8.48} & \good{15.66} & \good{1.3425} & \good{-7.98} & 10.70 & \good{1.1902} & \good{-8.21} & \good{16.26} & \best{1.6012} & -9.53 \\
\hline
\rowcolor[HTML]{D0DCE6} \multicolumn{13}{l}{\textbf{Agent Methods with  Gemini 3.1 Pro}} \\
\hline
TradingAgents & 13.76 & 1.0565 & -12.23 & 14.64 & 1.2229 & -11.20 & 3.72 & 0.3021 & -11.49 & 14.11 & 1.1851 & -12.44 \\
TradingGroup & 12.56 & 0.9666 & -12.82 & 13.22 & 1.0091 & -9.67 & 2.93 & 0.2752 & -10.22 & 9.89 & 0.9394 & -11.70 \\
RD-Agent & 15.42 & 1.3632 & -8.96 & 12.35 & 1.0902 & -10.55 & 7.27 & 0.8389 & -8.33 & 12.55 & 1.2801 & -9.20 \\
AlphaAgent & 16.92 & 1.3932 & -10.36 & 12.09 & 1.1938 & -10.81 & 7.86 & 1.0022 & -10.96 & 14.11 & 1.3372 & -11.05 \\
AlphaCrafter & 17.22 & 1.4852 & -9.27 & 14.52 & 1.3126 & -8.65 & 9.91 & \best{1.2001} & \best{-8.17} & 14.25 & 1.4008 & \best{-8.89} \\
\hline
\end{tabular*}
\end{table}

Traditional technical indicators, particularly MACD, achieve substantial raw returns by capturing market beta, yet suffer from unfavorable risk-adjusted profiles characterized by low Sharpe ratios and elevated drawdowns, reflecting the absence of effective risk management mechanisms. Machine learning and deep learning methods, notably LSTM, demonstrate strong backtesting performance but exhibit pronounced degradation in live trading, exposing their vulnerability to overfitting and regime shifts. This pattern is consistent across both markets, underscoring the fundamental challenge of generalizing data-driven models to out-of-sample conditions. Agent-based methods reveal a clear dichotomy. Role-playing agents (TradingAgents, TradingGroup) exhibit considerable cross-model instability due to the inherent variability in LLM reasoning across different backbone models. In contrast, quantitative trading agents (RD-Agent, AlphaAgent, AlphaCrafter) demonstrate more consistent behavior, attributed to their structured factor-centric workflows. Across the quantitative trading agent category, AlphaCrafter consistently exhibits favorable cross-model and cross-trial stability throughout both backtesting and live trading phases, suggesting that its harness-driven design contributes to more reliable and reproducible performance under varied conditions.

\subsection{Stability Analysis}
\label{subsec:stability_analysis}

\begin{figure}[htbp]
\centering
\begin{minipage}{0.48\textwidth}
    \centering
    \includegraphics[width=\textwidth]{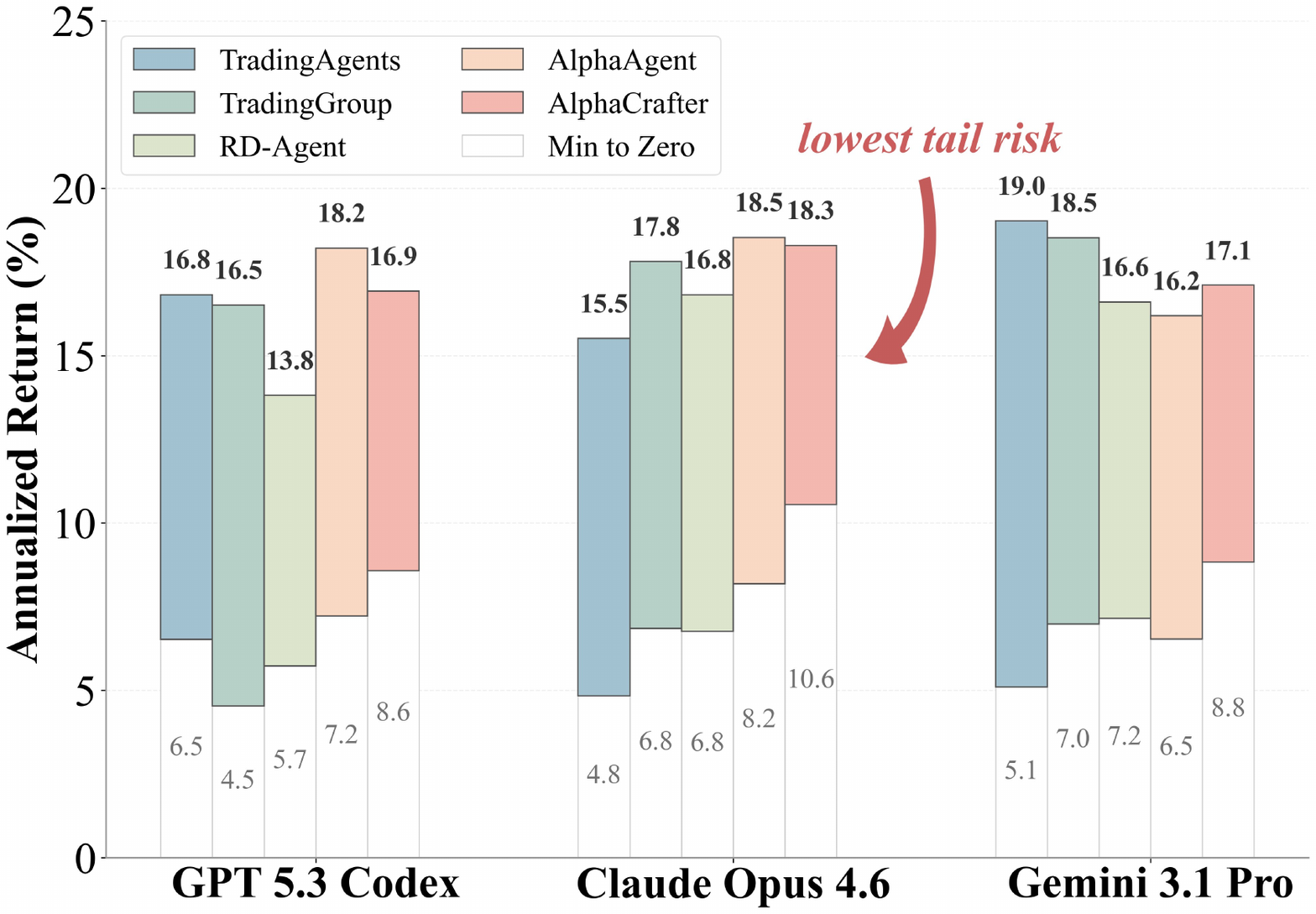}
    \caption{Performance distributions across independent backtesting trials. Each bar represents the min-max range of annualized returns across 10 independent trials for each method.}
    \label{fig:stability_trial}
\end{minipage}
\hfill
\begin{minipage}{0.48\textwidth}
    \centering
    \includegraphics[width=\textwidth]{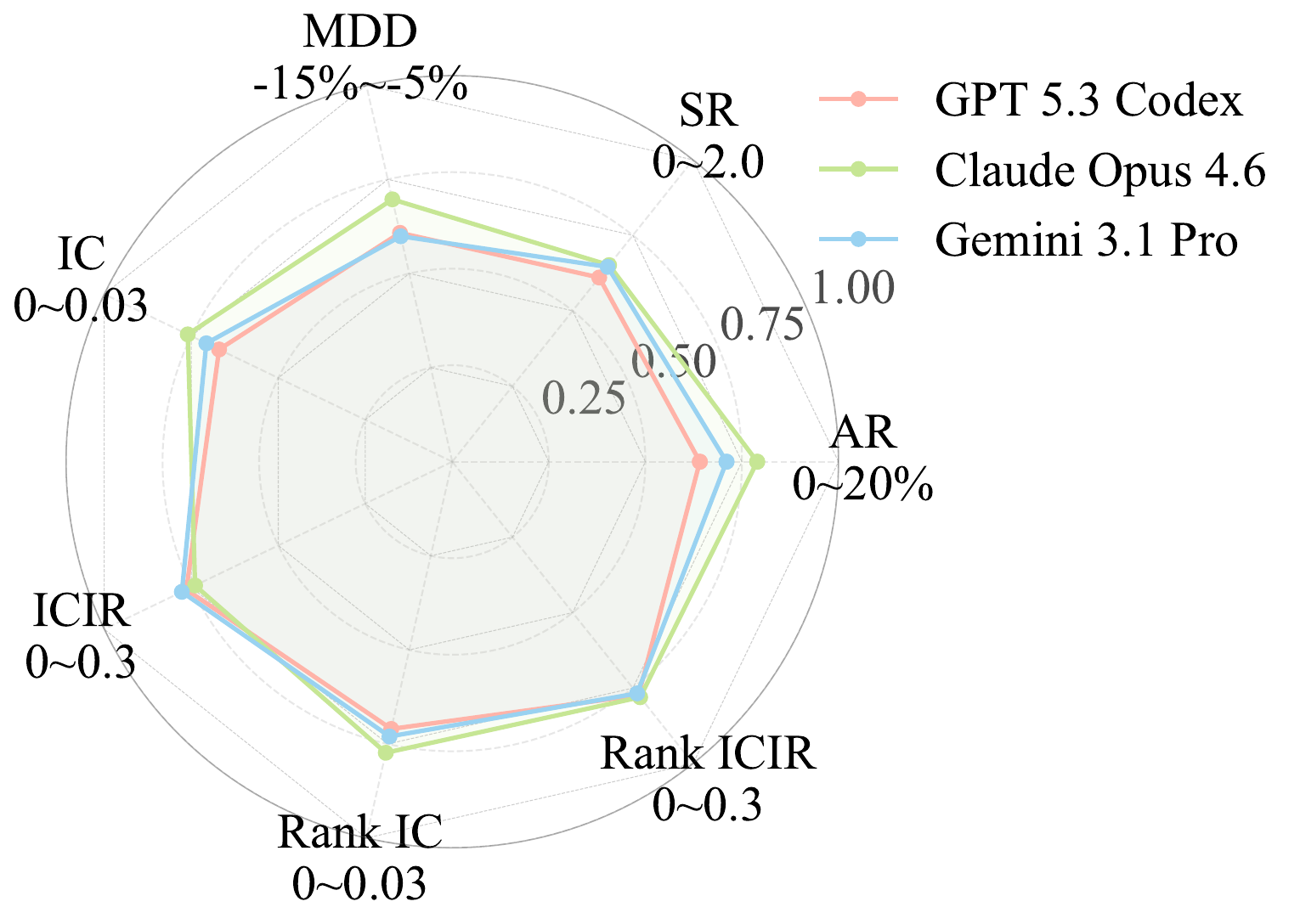}
    \caption{Performance comparison across backbone LLMs in backtesting. Radar chart comparing key metrics of AlphaCrafter instantiated with three backbone LLMs.}
    \label{fig:stability_model}
\end{minipage}
\end{figure}

To assess the reliability and robustness of AlphaCrafter, we aggregate all S\&P 500 backtesting experiments and conduct stability analyses from two perspectives: (1) performance consistency under stochastic variation through repeated independent trials, and (2) model robustness with respect to the choice of underlying LLM backbone. Results are presented in Figure~\ref{fig:stability_trial} and Figure~\ref{fig:stability_model}.

\subsubsection{Trial Stability}

Figure~\ref{fig:stability_trial} presents the distribution of annualized returns across 10 independent backtesting trials. AlphaCrafter exhibits the smallest cross-trial variance and the highest minimum returns among all methods, indicating the lowest tail risk. This robustness stems from the structured harness mechanism, which constrains agent behavior through programmable policies rather than unconstrained prompt execution. In contrast, baseline methods display wider variance and pronounced negative outliers, confirming that AlphaCrafter delivers reliable and reproducible performance under repeated evaluation.

\subsubsection{Model Stability}

Figure~\ref{fig:stability_model} compares AlphaCrafter's backtesting performance across three backbone LLMs. The framework exhibits a consistently stable profile across all models, with radar patterns remaining broadly aligned across all metric dimensions. Notably, AlphaCrafter demonstrates the smallest cross-model range across all performance metrics, confirming that the harness mechanism effectively decouples the framework's effectiveness from the specific choice of underlying LLM. While Claude Opus 4.6 yields marginally superior overall performance, the GPT and Gemini variants maintain robust results with minimal degradation. This stability underscores the model-agnostic design of AlphaCrafter, where procedural workflows and verification mechanisms embedded in the agent harness ensure consistent behavior across diverse model architectures, a critical property for practical deployment where LLM choices may vary due to cost or availability constraints.

\subsection{Alpha Decay Analysis}
\label{sec:alpha_decay_analysis}

Alpha decay refers to the gradual erosion of a factor's predictive power over time, as market dynamics shift and the informational edge embedded in the factor becomes arbitraged away by other market participants. To evaluate the temporal stability of factor efficacy and the extent to which different curation strategies mitigate this decay, we conduct an analysis across four consecutive semi-annual periods from January 2024 to January 2026. We report the mean, maximum, and minimum Information Coefficient for agent-based quantitative trading methods (RD-Agent, AlphaAgent, and AlphaCrafter), alongside two Alpha158-based baselines \cite{yang2020qlib}: \textbf{global top20} (retains the 20 best factors over the entire backtesting horizon) and \textbf{periodic top20} (dynamically re-selects the top 20 factors within each semi-annual interval). For AlphaCrafter, we dynamically track the effective factors retained in the factor library $\mathcal{Z}_t$, reflecting the system's adaptive curation process. All agent-based results reported in this analysis are instantiated with the Claude Opus 4.6 backbone, which yields the best overall performance in the main experiments.

\begin{figure}[htbp]
\centering
\begin{subfigure}{0.48\textwidth}
    \centering
    \includegraphics[width=\textwidth]{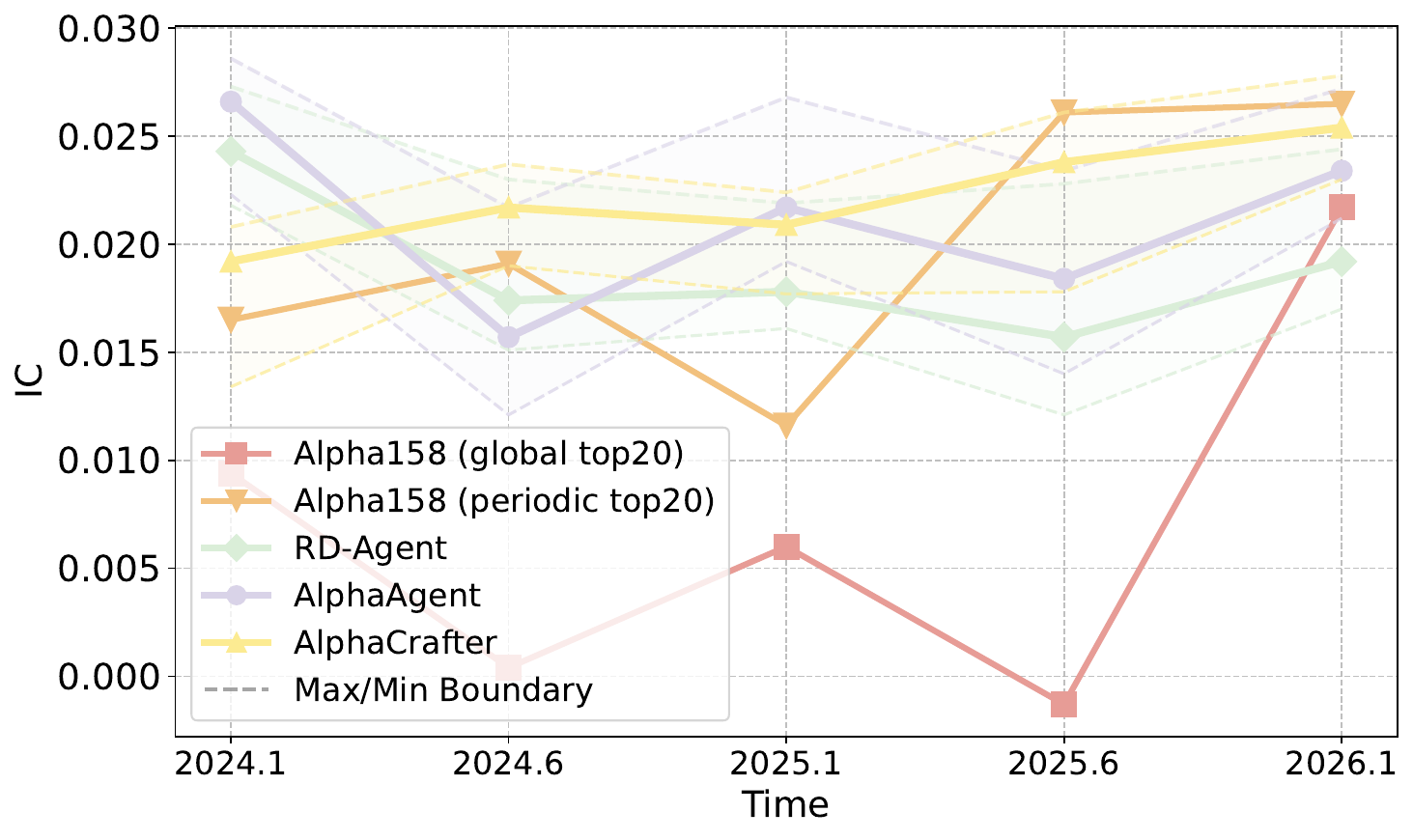}
    \caption{CSI 300 Market}
    \label{fig:alpha_decay_csi300}
\end{subfigure}
\hfill
\begin{subfigure}{0.48\textwidth}
    \centering
    \includegraphics[width=\textwidth]{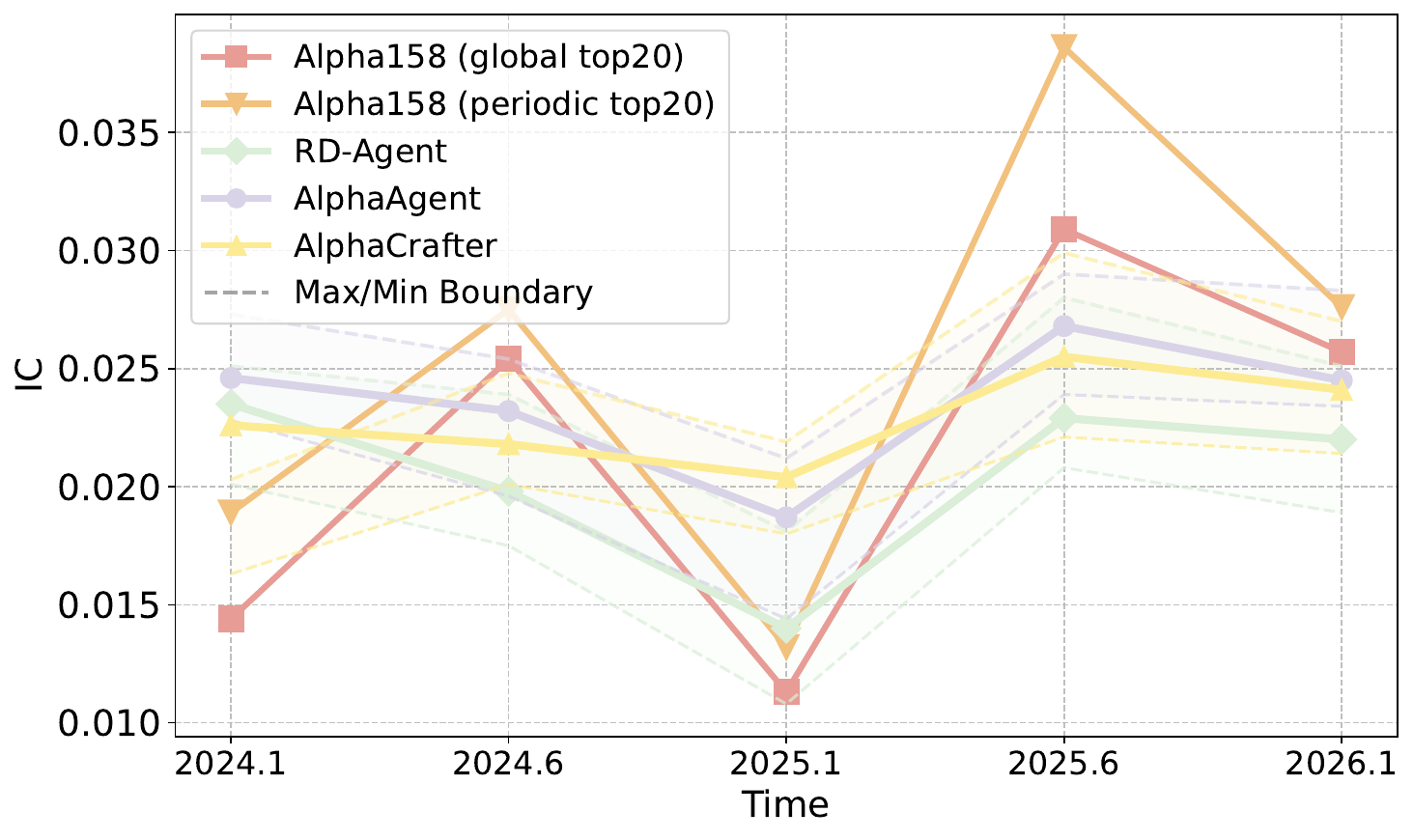}
    \caption{S\&P 500 Market}
    \label{fig:alpha_decay_sp500}
\end{subfigure}
\caption{IC comparison of different methods across time periods on CSI 300 and S\&P 500 markets.}
\label{fig:alpha_decay_analysis}
\end{figure}

As shown in Figure~\ref{fig:alpha_decay_analysis}, the two Alpha158 baselines exhibit diametrically opposite behaviors: \textit{periodic top20} maintains consistently high IC across both markets through frequent refreshment, while \textit{global top20} suffers severe volatility with IC values fluctuating widely and even turning negative, confirming that static factor sets are highly vulnerable to market regime shifts. Among agent-based methods, AlphaAgent and AlphaCrafter sustain stable IC levels around 0.02 across all periods, demonstrating robust predictive persistence. In contrast, RD-Agent exhibits a noticeable declining trend in later evaluation windows, indicating more pronounced alpha decay. These results collectively demonstrate that both regularization-based search and adaptive library maintenance are effective in mitigating alpha decay, underscoring the importance of dynamic factor curation for preserving predictive power in evolving markets.

\subsection{Ablation Study}
\label{subsec:ablation}

To quantify the marginal contribution of each component, we conduct ablation experiments with the following variants: (1) \textbf{w/o Miner}, replacing the Miner Cluster harness with the static Alpha158 factor set \cite{yang2020qlib}; (2) \textbf{w/o Screener}, replacing the Screener harness with equal-weighted allocation across all available factors in $\mathcal{Z}_t$; (3) \textbf{w/o Trader}, replacing the Trader harness with the fixed reference strategy $\pi_{\text{ref}}$; (4) \textbf{w/o Harness}, retaining all specialized agents but replacing the programmatic harness with a natural-language workflow in which the agents coordinate through natural-language instructions; and (5) \textbf{Single-Agent}, using a single LLM agent that performs all programmatic workflows within a single context, bypassing the multi-agent harness design. All experiments use Claude Opus 4.6 as the backbone model.

\begin{table}[htbp]
\centering
\caption{Ablation Study: Backtesting Performance on CSI 300 and S\&P 500}
\label{tab:ablation}
\small
\setlength{\tabcolsep}{2.7pt}
\begin{tabular}{l|cccc|cccc}
\hline
\multirow{2}{*}{\textbf{Method}} & \multicolumn{4}{c|}{\textbf{CSI 300}} & \multicolumn{4}{c}{\textbf{S\&P 500}} \\
& \textbf{AR(\%)} ↑ & \textbf{SR} ↑ &\textbf{MDD(\%)} ↑ & \textbf{AR(\%) Std} ↓ & \textbf{AR(\%)} ↑ & \textbf{SR} ↑ & \textbf{MDD(\%)} ↑ & \textbf{AR(\%) Std} ↓ \\
\hline
\multicolumn{9}{l}{\textbf{Ablation Variants}} \\
\hline
w/o Miner & 16.64 & 1.3515 & -9.82 & 3.53 & 14.02 & 1.1401 & -9.26 & 3.28 \\
w/o Screener & 15.87 & 1.3858 & -9.24 & 3.96 & 13.73 & 1.1377 & -10.42 & 4.12 \\
w/o Trader & 16.11 & 1.3284 & -10.85 & 4.15 & 13.32 & 1.1250 & -11.15 & 3.75 \\
w/o Harness & 15.32 & 1.3628 & -10.93 & 4.02 & 14.53 & 1.2769 & -8.58 & 3.31 \\
Single-Agent & 14.83 & 1.1527 & -12.24 & 4.57 & 9.95 & 0.8631 & -11.37 & 4.16 \\
\hline
\multicolumn{9}{l}{\textbf{Full Model}} \\
\hline
\rowcolor[HTML]{E9F7EA} AlphaCrafter & \best{18.88} & \best{1.6732} & \best{-8.48} & \best{3.03} & \best{15.66} & \best{1.3425} & \best{-7.98} & \best{2.91} \\
\hline
\end{tabular}
\end{table}

Table~\ref{tab:ablation} reports the results. Removing any component-level harness degrades performance across all metrics. The w/o Miner variant, which replaces the Miner Cluster harness, exhibits notable AR decline on both markets, as the static Alpha158 factor set suffers from IC instability and alpha decay, whereas the Miner Cluster harness continuously generates and validates new factors. The w/o Screener variant, which removes the Screener harness, incurs substantial AR degradation, confirming its role in selecting regime-appropriate factors and mitigating downside risk. The w/o Trader variant, which replaces the Trader harness, yields low Sharpe ratios, showing that adaptive execution is essential for risk-adjusted returns.

The w/o Harness variant holds the agent set fixed but replaces structured programmatic interfaces with natural-language coordination. On S\&P 500 it drops AR from 15.66 to 14.53 and SR from 1.3425 to 1.2769 while raising AR Std from 2.91 to 3.31; CSI 300 shows a consistent degradation. Since the agent decomposition is unchanged, this gap isolates the stabilizing role of the programmatic harness: structured interfaces constrain and validate each step of agent interaction, which natural-language coordination alone does not provide.

The Single-Agent variant exhibits the highest variance and most severe drawdowns on both markets, as the monolithic agent must manage factor generation, regime diagnosis, ensemble selection, and strategy construction within a single context window, causing critical steps to be lost or corrupted. The multi-agent harness instead provides explicit context isolation, and the full model achieves the highest AR, Sharpe ratio, and lowest volatility across both universes.

\section{Related Work}
\label{sec:related-work}

\subsection{LLM-Powered Agentic Systems}

LLM-based agentic systems span a broad spectrum, ranging from predefined workflows \cite{Zhuge_2025, xi2023risepotentiallargelanguage} to fully autonomous agents \cite{Wang_2024, zhang2024mobileexpertsdynamictoolenabledagent, hong2024datainterpreterllmagent}. Recent advances have shifted toward skill-based frameworks that treat agent capabilities as composable and verifiable modules, enabling dynamic assembly and reuse \cite{liang2026skillnetcreateevaluateconnect, xu2026agentskillslargelanguage, li2026skillsbenchbenchmarkingagentskills}. A growing body of evidence suggests that the harness surrounding an LLM—the structured interface that governs its behavior—is as critical as the model itself for sustained autonomous operation \cite{openai2026harness, young2025effective}. These insights motivate the design of agent systems with explicit behavioral constraints and verification mechanisms.

\subsection{LLM for Financial Trading}

Domain-specific financial LLMs such as FinGPT \cite{yang2025fingptopensourcefinanciallarge} and FinLlama \cite{konstantinidis2024finllamafinancialsentimentclassification} demonstrate strong performance on financial tasks such as sentiment analysis and information extraction, though they typically operate on static data without real-time adaptation. Agent-based trading architectures have emerged along several trajectories. News-driven agents incorporate sentiment signals from financial news and social media to inform trading decisions \cite{zhang2024unveilingpotentialsentimentlarge, wang2024llmfactorextractingprofitablefactors, yu2023finmemperformanceenhancedllmtrading}. Reflection-driven agents iteratively refine their decisions through self-critique and multi-agent debate \cite{Xing_2025, zhang2024aimeetsfinancestockagent, yu2024finconsynthesizedllmmultiagent, xiao2025tradingagentsmultiagentsllmfinancial, tian2025tradinggroupmultiagenttradingselfreflection}. Factor optimization frameworks leverage LLMs for automated alpha discovery and strategy formulation \cite{kou2025automatestrategyfindingllm, li2025rdagentquantmultiagentframeworkdatacentric, tang2025alphaagentllmdrivenalphamining, wang2024quantagentseekingholygrail, wang2025alphagpthumanaiinteractivealpha}. These approaches collectively demonstrate the potential of LLMs in automating various aspects of quantitative trading, from signal generation to decision-making.

\section{Conclusion}
\label{sec:conclusion}

In this paper, we introduced \textsc{AlphaCrafter}, a multi-agent framework grounded in harness engineering principles that formalizes quantitative trading workflows through structured algorithmic policies. Unlike existing agent-based systems that rely on loosely specified natural-language prompts, AlphaCrafter encapsulates each agent within programmable policy specifications that integrate procedural workflows, execution constraints, and explicit verification mechanisms. This harness-driven design transforms the entire trading pipeline into a sequence of well-defined, reproducible, and auditable decision processes with explicit execution semantics. Extensive evaluations on the CSI 300 and S\&P 500 benchmarks demonstrate that AlphaCrafter consistently achieves superior risk-adjusted returns while exhibiting substantially lower cross-model and cross-trial variance compared to state-of-the-art baselines. These results suggest that disciplined harness-based design provides a practical foundation for building more reliable, controllable, and robust multi-agent systems for quantitative trading, with clear potential for extension to broader financial decision-making domains.

\bibliographystyle{assets/plainnat}
\bibliography{references}

\clearpage
\beginappendix

\appendix
\section{Reference Strategy}
\label{sec:reference_strategy}

Algorithm~4 defines the fixed reference strategy $\pi_{\text{ref}}$, which serves as the non-adaptive baseline for the Trader agent. Given a set of factor signals from the ensemble $\mathcal{E}_t$, it computes a composite score $\phi_{i,t}$ for each asset as a weighted combination of factor values. Assets are then ranked by this score, with the top $N_{\text{long}}$ selected for long positions and the bottom $N_{\text{short}}$ for short positions. Position sizes are determined by a gross exposure parameter $\beta$ and a net exposure bias $\gamma$, which controls the long–short balance (e.g., $\gamma=1$ yields a long-only portfolio). The strategy submits orders to adjust holdings from the previous period to the computed target weights.

\begin{algorithm}[h]
\small
\caption{Reference Strategy $\pi_{\text{ref}}$}
\label{alg:reference_strategy}
\begin{algorithmic}[1]
\STATE $\Theta \gets \{\beta, \gamma, N_{\text{long}}, N_{\text{short}}\}$ \COMMENT{Hyperparameters}
\STATE $\text{NAV}_t \gets \text{net asset value at time } t$

\FOR{each $i \in \mathcal{U}$}
    \STATE $\phi_{i,t} \gets \sum_{j=1}^{|\mathcal{E}_t|} w_j \cdot d_j \cdot f_j(\mathbf{x}_{i,t-\ell+1:t})$ \COMMENT{Composite score}
\ENDFOR

\STATE $\text{rank} \gets \text{sort}(\mathcal{U}, \phi_{i,t}, \text{descending})$
\STATE $\mathcal{I}_{\text{long}} \gets \text{top}(\text{rank}, N_{\text{long}}),\;\; \mathcal{I}_{\text{short}} \gets \text{bottom}(\text{rank}, N_{\text{short}})$

\FOR{each $i \in \{i \mid h_{i,t-1} \neq 0\}$}
    \IF{$i \notin \mathcal{I}_{\text{long}} \cup \mathcal{I}_{\text{short}}$}
        \STATE $\text{submit\_order}(i, -h_{i,t-1})$ \COMMENT{Liquidate exited positions}
    \ENDIF
\ENDFOR

\STATE $V_{\text{long}} \gets \beta \cdot \text{NAV}_t \cdot (1 + \gamma) / 2$
\STATE $V_{\text{short}} \gets \beta \cdot \text{NAV}_t \cdot (1 - \gamma) / 2$

\FOR{each $i \in \mathcal{I}_{\text{long}}$}
    \STATE $h_{i,t}^{\text{target}} \gets V_{\text{long}} / (N_{\text{long}} \cdot P_{i,t})$
\ENDFOR

\FOR{each $i \in \mathcal{I}_{\text{short}}$}
    \STATE $h_{i,t}^{\text{target}} \gets -V_{\text{short}} / (N_{\text{short}} \cdot P_{i,t})$
\ENDFOR

\FOR{each $i \in \mathcal{I}_{\text{long}} \cup \mathcal{I}_{\text{short}}$}
    \STATE $\text{submit\_order}(i, h_{i,t}^{\text{target}} - h_{i,t-1})$ \COMMENT{Adjust to target holdings}
\ENDFOR
\end{algorithmic}
\end{algorithm}

\section{Experimental Details}
\label{sec:experimental_details}

\subsection{Dataset Details}
\label{subsec:dataset_details}

\subsubsection{Data Sources}
The daily OHLCV data for CSI 300 constituents is collected from \textbf{Baostock} \cite{baostock2026}. For the S\&P 500 constituents, daily price-volume data is obtained from \textbf{Yahoo Finance} \cite{aroussi2026yfinance}. Fundamental indicators (PE, PS, PB, DYR), financial statement data (quarterly balance sheets, income statements, cash flow statements), and alternative data (financial news and corporate announcements) are sourced from \textbf{Lixinger} \cite{lixinger2026}.

\subsubsection{Data Storage Format}
Data are stored in the following formats:
\begin{itemize}
    \item \textbf{Daily OHLCV data:} Stored in CSV format, with each row corresponding to a trading day and columns representing Open, High, Low, Close, and Volume.
    \item \textbf{Fundamental indicators:} Stored in CSV format, with each row corresponding to a trading day per asset and columns representing PE, PS, PB, and DYR.
    \item \textbf{Financial statements:} Stored in JSON format, organized hierarchically by asset ticker and reporting quarter. Each JSON object contains standardized fields for balance sheet, income statement, and cash flow statement items.
    \item \textbf{Alternative data (news and announcements):} Stored in JSON format, where each entry contains the publication timestamp, asset ticker, headline, full content, and sentiment metadata (when available).
\end{itemize}

\subsection{Metrics Details}
\label{subsec:metrics_details}

We provide the mathematical formulations for the evaluation metrics used throughout our framework. The notation follows the conventions established in Section~\ref{subsec:miner_cluster} and Section~\ref{subsec:screener_agent}.

\paragraph{Annualized Return (AR).}
Given a sequence of daily portfolio values $V_0, V_1, \dots, V_T$ over an evaluation period of $T$ trading days, the total return is $R_{\text{total}} = \frac{V_T - V_0}{V_0}$. The Annualized Return is computed as:
\begin{equation}
\text{AR} = \left(1 + R_{\text{total}}\right)^{\frac{D}{T}} - 1
\end{equation}
where $D$ denotes the number of trading days per calendar year, with $D = 243$ for the CSI 300 (Chinese A-share market) and $D = 252$ for the S\&P 500 (U.S. equity market).

\paragraph{Sharpe Ratio (SR).}
Let $r_t = \frac{V_t - V_{t-1}}{V_{t-1}}$ denote the daily portfolio return on day $t$. The Sharpe Ratio is defined as:
\begin{equation}
\text{SR} = \frac{\sqrt{D} \cdot \left(\frac{1}{T}\sum_{t=1}^{T} r_t - r_f\right)}{\sqrt{\frac{1}{T-1}\sum_{t=1}^{T}(r_t - \bar{r})^2}}
\end{equation}
where $D$ is the number of trading days per year, $r_f$ is the daily risk-free rate. Following conventional practice, we set the annualized risk-free rate to 1.25\% for the Chinese market (CSI 300) and 3.81\% for the U.S. market (S\&P 500), based on the 2-year government bond yields. Source: \textbf{Trading Economics} \cite{tradingeconomics_2026}.

\paragraph{Maximum Drawdown (MDD).}
Let $V_t$ denote the portfolio value at time $t$. The Maximum Drawdown is the largest peak-to-trough decline over the evaluation period:
\begin{equation}
\text{MDD} = \min_{t \in [0,T]} \left( \frac{V_t - \max_{s \in [0,t]} V_s}{\max_{s \in [0,t]} V_s} \right)
\end{equation}

\paragraph{Information Coefficient (IC).}
For a given factor $f$ and a cross-section of $N$ assets at time $t$, let $\mathbf{f}_t \in \mathbb{R}^N$ denote the vector of factor signals:
\begin{equation}
\mathbf{f}_t = \bigl(f(\mathbf{x}_{1,t-\ell+1:t}), \ldots, f(\mathbf{x}_{N,t-\ell+1:t})\bigr)^\top \in \mathbb{R}^N
\end{equation}
and let $\mathbf{r}_{t+h} \in \mathbb{R}^N$ denote the vector of forward returns from time $t$ to $t+h$:
\begin{equation}
\mathbf{r}_{t+h} = (r_{1,t+h}, \ldots, r_{N,t+h})^\top \in \mathbb{R}^N
\end{equation}
where $h$ denotes the prediction horizon. The IC at time $t$ for horizon $h$ is the Pearson correlation coefficient between the factor signal and subsequent returns:
\begin{equation}
\text{IC}_t^{(h)} = \text{corr}(\mathbf{f}_t, \mathbf{r}_{t+h})
\end{equation}
Similarly, the Rank IC at time $t$ for horizon $h$ is the Spearman rank correlation:
\begin{equation}
\text{RankIC}_t^{(h)} = \text{rank\_corr}(\mathbf{f}_t, \mathbf{r}_{t+h})
\end{equation}
In our empirical experiments, all reported IC and RankIC results are these time-averaged values with $h=1$ (i.e., next-day returns) unless otherwise specified.

\paragraph{Information Coefficient Information Ratio (ICIR).}
Given a time series of IC values $\{\text{IC}_t\}_{t=1}^{T}$, the ICIR is defined as:
\begin{equation}
\text{ICIR} = \frac{\overline{\text{IC}}}{\sigma_{\text{IC}}}
\end{equation}
where $\overline{\text{IC}} = \frac{1}{T}\sum_{t=1}^{T} \text{IC}_t$ is the mean IC, and $\sigma_{\text{IC}} = \sqrt{\frac{1}{T-1}\sum_{t=1}^{T}(\text{IC}_t - \overline{\text{IC}})^2}$ is the standard deviation of IC. The Rank ICIR is computed analogously:
\begin{equation}
\text{RankICIR} = \frac{\overline{\text{RankIC}}}{\sigma_{\text{RankIC}}}
\end{equation}

\paragraph{Hit Ratio.}
The Hit Ratio measures the directional accuracy of the factor's predictions. It is defined as the proportion of periods in which the IC is positive, indicating that the factor correctly predicted the direction of relative asset performance:
\begin{equation}
\eta = \frac{1}{T} \sum_{t=1}^{T} \mathbf{1}\left[\psi_t > 0\right]
\end{equation}
where $\mathbf{1}[\cdot]$ denotes the indicator function.

\paragraph{Coverage.}
Coverage measures the proportion of assets for which the factor produces a valid (non-NA) signal at time $t$:
\begin{equation}
\kappa_t = \frac{\left|\left\{i \in [N] : f(\mathbf{x}_{i,t-\ell+1:t}) \neq \text{NA}\right\}\right|}{N}
\end{equation}
The overall coverage is averaged across all time periods:
\begin{equation}
\kappa = \frac{1}{T}\sum_{t=1}^{T} \kappa_t
\end{equation}

\paragraph{Turnover.}
Turnover quantifies the stability of the factor's ranking across consecutive periods. At time $t$, it is computed as the average absolute rank change normalized by the number of assets:
\begin{equation}
\tau_t = \frac{\|\text{rank}(\mathbf{f}_t) - \text{rank}(\mathbf{f}_{t-1})\|_1}{N}
\end{equation}
where $\text{rank}(\mathbf{f}_t) \in \mathbb{R}^N$ denotes the vector of ranks assigned to each asset based on factor values, and $\|\cdot\|_1$ is the $L_1$ norm. The average turnover is given by:
\begin{equation}
\tau = \frac{1}{T-1}\sum_{t=2}^{T} \tau_t
\end{equation}

\subsection{Simulation Settings}
\label{subsec:simulation_settings}

To provide a realistic yet tractable evaluation environment, we implement market-specific backtesting simulators that approximate the trading rules of the Chinese A-share market (CSI 300) and the U.S. equity market (S\&P 500). These simulators model essential exchange mechanisms, including order execution, account management, and transaction costs, while intentionally simplifying certain aspects of real-world market microstructure. The key simulation settings are summarized below.

\textbf{CSI 300 Exchange (A-share Market).}
The CSI 300 simulator is configured to approximate the trading constraints of the Chinese A-share market. In particular, it models the T+1 settlement rule by preventing shares purchased on a trading day from being sold until the following trading day. Since short selling is generally inaccessible in typical retail-oriented trading scenarios, the simulator adopts a long-only setting. A proportional transaction cost of $0.02\%$ (2 basis points) is applied to each trade as a simplified approximation of brokerage commissions and related execution costs.

\textbf{S\&P 500 Exchange (U.S. Market).}
The S\&P 500 simulator approximates the trading conventions of the U.S. equity market by allowing same-day position adjustments and supporting long-short portfolio construction. To model margin constraints in a simplified manner, opening short positions assumes an initial margin requirement of $50\%$ of the position value, while a maintenance margin of $30\%$ is used throughout the backtesting process. Rather than reproducing the fee schedule of any specific broker, the simulator applies a fixed proportional transaction cost of $0.01\%$ (1 basis point) per trade to account for commissions and other execution-related costs.

\textbf{Integration with Reference Strategy.}
These market-specific simulation settings are naturally integrated with the reference strategy $\pi_{\text{ref}}$ (Algorithm~4). The hyperparameter $\gamma \in [-1,1]$ controls the net exposure bias. For the CSI 300 simulator, where only long positions are considered, we set $\gamma = 1$. For the S\&P 500 simulator, which supports long-short portfolios, we set $\gamma = 0.5$ to maintain a moderate net long bias while permitting partial short exposure. Across both simulated markets, the gross exposure parameter is fixed at $\beta = 0.8$, corresponding to approximately $80\%$ capital deployment and leaving the remaining capital as cash to accommodate execution uncertainty and simplified margin requirements during backtesting.

\section{Case Study}
\label{sec:case_study}

To complement the quantitative results, we conduct a series of case studies examining the internal behavior of AlphaCrafter's harness-driven pipeline. Specifically, we investigate three aspects: (1) whether the Miner harness generates genuinely novel factors rather than syntactic recombinations of existing ones, (2) whether the Screener harness accurately perceives market regimes through its structured information retrieval and LLM-based assessment, and (3) whether the integrated harness system exhibits coherent risk-awareness through adaptive position sizing.

\subsection{Factor Semantic Diversity and Novelty Analysis}

To assess whether factors discovered by AlphaCrafter's Miner harness constitute genuinely novel alpha signals rather than syntactic recombinations of classical factors, we conduct a semantic diversity analysis grounded in operator-tree topology.

Every factor expression is parsed into a normalized abstract syntax tree (AST), where leaf nodes represent raw data (e.g., \texttt{close}, \texttt{volume}) and internal nodes denote operators (e.g., \texttt{rank}, \texttt{ts\_corr}). Let $\delta(f_i, f_j) \in [0,1]$ denote the normalized tree edit distance (NTED) between two factors, computed by dividing the raw edit distance by the sum of the two tree sizes. This metric captures structural divergence at the algorithmic level, independent of parameter values such as rolling-window lengths.

We define two complementary indices. For a factor library $\mathcal{A}^{(k)}$ produced in trial $k$, let $n_k = |\mathcal{A}^{(k)}|$ be the number of factors in that trial, and let $a_i^{(k)} \in \mathcal{A}^{(k)}$ denote the $i$-th factor expression. The \textbf{internal semantic diversity} is
\begin{equation}
\Phi_{\text{intra}}^{(k)} = \frac{2}{n_k(n_k-1)} \sum_{1 \le i < j \le n_k} \delta\big(a_i^{(k)},\; a_j^{(k)}\big),
\end{equation}
which computes the arithmetic mean of the NTED over all unordered pairs of distinct factors within the same trial. A high $\Phi_{\text{intra}}$ indicates that the Miner explores a broad logical spectrum rather than resampling a narrow set of templates.

Let $\mathcal{B}$ denote the Alpha158 \cite{qlib2026alpha158} classical reference library, with $b_j \in \mathcal{B}$ indexing its constituent factors. The \textbf{external semantic novelty} against $\mathcal{B}$ is
\begin{equation}
\Phi_{\text{inter}}^{(k)} = \frac{1}{n_k} \sum_{i=1}^{n_k} \min_{b_j \in \mathcal{B}} \; \delta\big(a_i^{(k)},\; b_j\big),
\end{equation}
where the inner minimization identifies, for each generated factor $a_i^{(k)}$, its nearest structural neighbour in the classical library. A high $\Phi_{\text{inter}}$ implies systematic departure from established factor structures.

For $K$ independent trials, we report the trial-averaged metrics
\begin{equation}
\bar{\Phi}_{\text{intra}} = \frac{1}{K} \sum_{k=1}^{K} \Phi_{\text{intra}}^{(k)}, \qquad
\bar{\Phi}_{\text{inter}} = \frac{1}{K} \sum_{k=1}^{K} \Phi_{\text{inter}}^{(k)}.
\end{equation}

\begin{figure}[htbp]
\centering
\begin{subfigure}{0.48\textwidth}
    \centering
    \includegraphics[width=\textwidth]{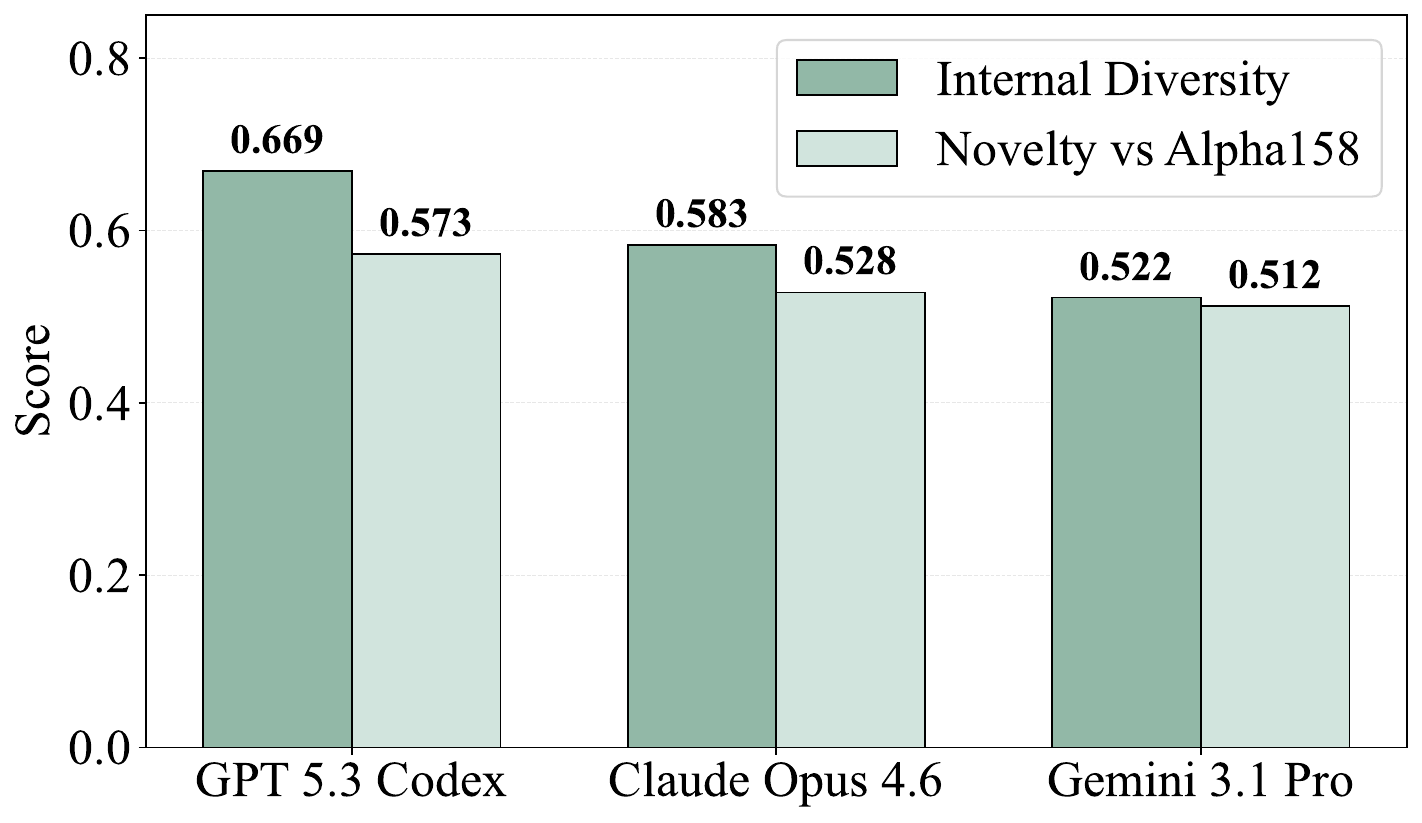}
    \caption{CSI~300 Market}
    \label{fig:factor_case_csi300}
\end{subfigure}
\hfill
\begin{subfigure}{0.48\textwidth}
    \centering
    \includegraphics[width=\textwidth]{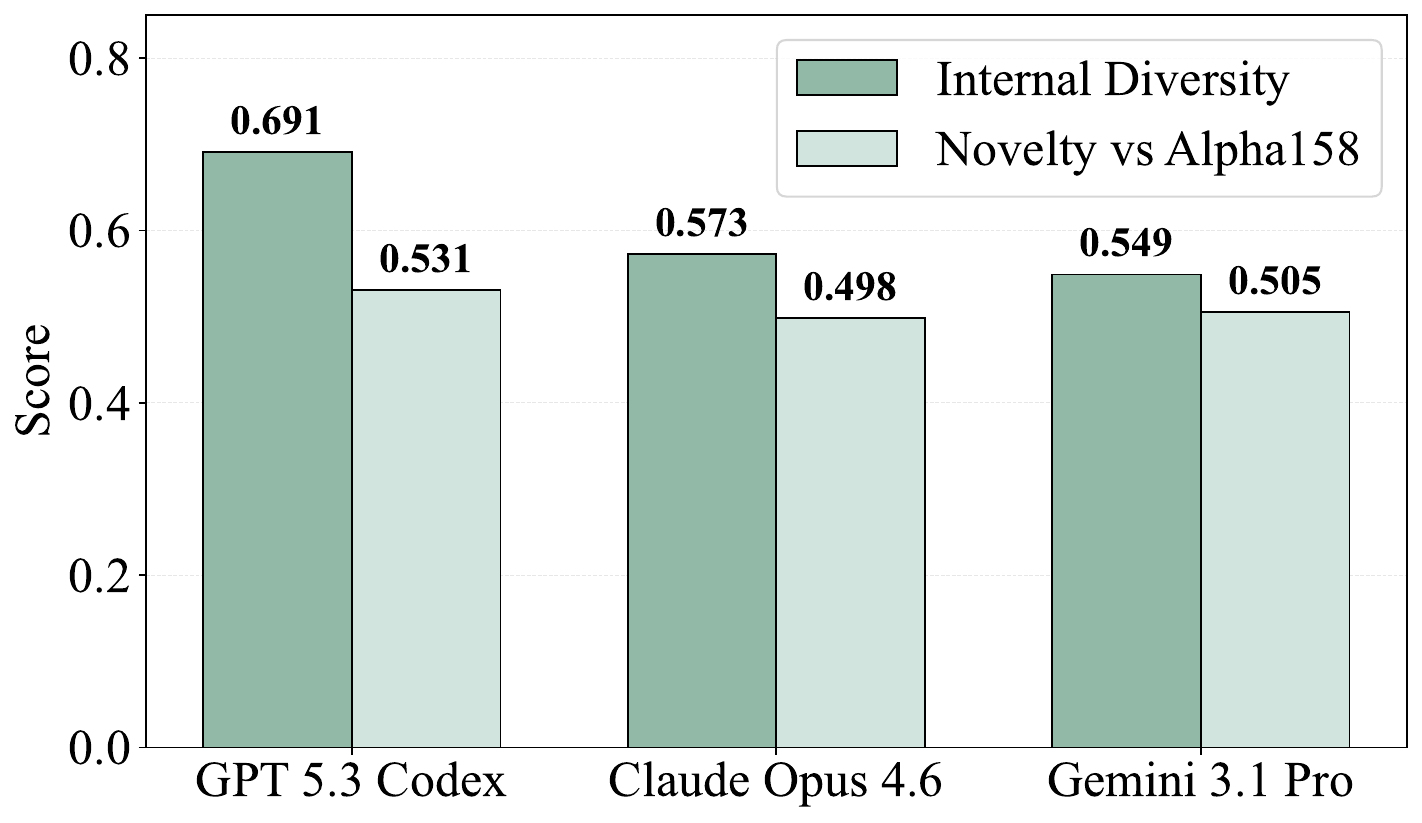}
    \caption{S\&P~500 Market}
    \label{fig:factor_case_sp500}
\end{subfigure}
\caption{Semantic diversity and novelty metrics across three LLM backbones.}
\label{fig:factor_case_study}
\end{figure}

Figure~\ref{fig:factor_case_study} reports $\bar{\Phi}_{\text{intra}}$ and $\bar{\Phi}_{\text{inter}}$ across both markets. Two principal findings emerge. GPT 5.3 Codex excels at structural novelty, consistently achieving the highest $\bar{\Phi}_{\text{inter}}$, indicating that it generates factors whose operator-tree configurations lie furthest from classical Alpha158 formulations. Its $\bar{\Phi}_{\text{intra}}$ also leads, suggesting that its exploratory mechanism spans diverse logical structures. Claude Opus 4.6 and Gemini 3.1 Pro exhibit moderate but stable novelty scores.

Critically, these semantic properties exhibit a nuanced relationship with downstream trading performance. Despite GPT 5.3 Codex's superiority in generating structurally novel factors, its realized risk-adjusted returns do not surpass those of Claude Opus 4.6. This divergence carries an important implication: \textbf{greater factor novelty does not monotonically translate into superior investment outcomes}. The Screener harness's role in filtering and ensemble construction under market-regime awareness appears to act as a moderating mechanism, selecting factors that are not merely original but also seasonally appropriate. This underscores the value of the harness architecture, where the Screener's structured verification complements the Miner's generative exploration.

\subsection{Regime Coherence Analysis}
\label{sec:regime_coherence}

To evaluate the fidelity of AlphaCrafter's market perception, we analyze the alignment between the Screener harness's semantic regime assessments and empirically measured market conditions. The Screener harness operates through a structured two-step process: it first retrieves quantitative market information via a dedicated tool, then invokes an LLM to produce qualitative regime diagnoses. This design ensures that assessments are grounded in actual data rather than relying solely on the LLM's parametric knowledge. We conduct this case study using a representative trial of AlphaCrafter powered by Claude Opus 4.6, selected as the median-performing run to avoid cherry-picking. For each trading cycle, the Screener produces qualitative judgments across three distinct market dimensions: trend direction, volatility regime, and correlation structure.

We map each semantic label to a discrete numerical value in the set $\{0, 0.25, 0.5, 0.75, 1\}$ based on its ordinal intensity. For instance, trend labels are mapped as: ``strong downtrend'' $\to 0$, ``downtrend'' $\to 0.25$, ``range-bound'' $\to 0.5$, ``uptrend'' $\to 0.75$, ``strong uptrend'' $\to 1$. Volatility and correlation labels follow analogous mappings from ``low'' to ``high'' and from ``low dispersion'' to ``index-led'', respectively.

For each cycle $c$, we compute quantitative market proxies using trailing windows ending at timestamp $\tau_c$, with $L=20$ days for volatility and correlation, and $L=60$ days for trend. Historical data preceding the evaluation period is used to pre-warm all metrics.

\textbf{Trend proxy} $M_c^{\text{(trend)}}$: the normalized cumulative return of the market index over the past 60 trading days, mapped to $[0,1]$ via a logistic transformation:
\begin{equation}
M_c^{\text{(trend)}} = \frac{1}{1 + e^{-r_c^{(60)} / \sigma_0}},
\end{equation}
where $r_c^{(60)}$ is the 60-day log return, and $\sigma_0 = \sigma_{\text{ann}} \times \sqrt{60/D}$ with $\sigma_{\text{ann}} = 0.2$ as the long-term annualized volatility estimate.

\textbf{Volatility proxy} $M_c^{\text{(vol)}}$: the realized volatility of the index over the past 20 trading days, min-max normalized to $[0,1]$ using the empirical 5th and 95th percentiles from the full sample:
\begin{equation}
M_c^{\text{(vol)}} = \min\left(\max\left(\frac{\sigma_c^{(20)} - q_{0.05}}{q_{0.95} - q_{0.05}}, 0\right), 1\right),
\end{equation}
where $\sigma_c^{(20)} = \sqrt{\frac{D}{19} \sum_{i=1}^{20} (r_{t-i+1} - \bar{r})^2}$ is the annualized realized volatility.

\textbf{Correlation proxy} $M_c^{\text{(corr)}}$: the average absolute pairwise correlation of daily returns among index constituents:
\begin{equation}
M_c^{\text{(corr)}} = \frac{2}{K(K-1)} \sum_{i=1}^{K-1} \sum_{j=i+1}^{K} |\rho_{ij,c}^{(20)}|,
\end{equation}
where $\rho_{ij,c}^{(20)}$ is the 20-day Pearson correlation between stocks $i$ and $j$, and $K$ is the number of constituents with sufficient trading history.

We evaluate the coherence between Screener's semantic assessments and market-derived proxies over 50 consecutive trading cycles. For each dimension, we construct a similarity matrix where cell $(i, j)$ represents the similarity between the semantic assessment at cycle $i$ and the market proxy at cycle $j$, defined as $1 - |\text{semantic}_i - \text{market}_j|$. The raw similarity values are then linearly normalized to $[0,1]$ across the entire matrix, with $1$ indicating perfect alignment and $0$ indicating maximal dissimilarity.

\begin{figure}[htbp]
    \centering
    \includegraphics[width=\textwidth]{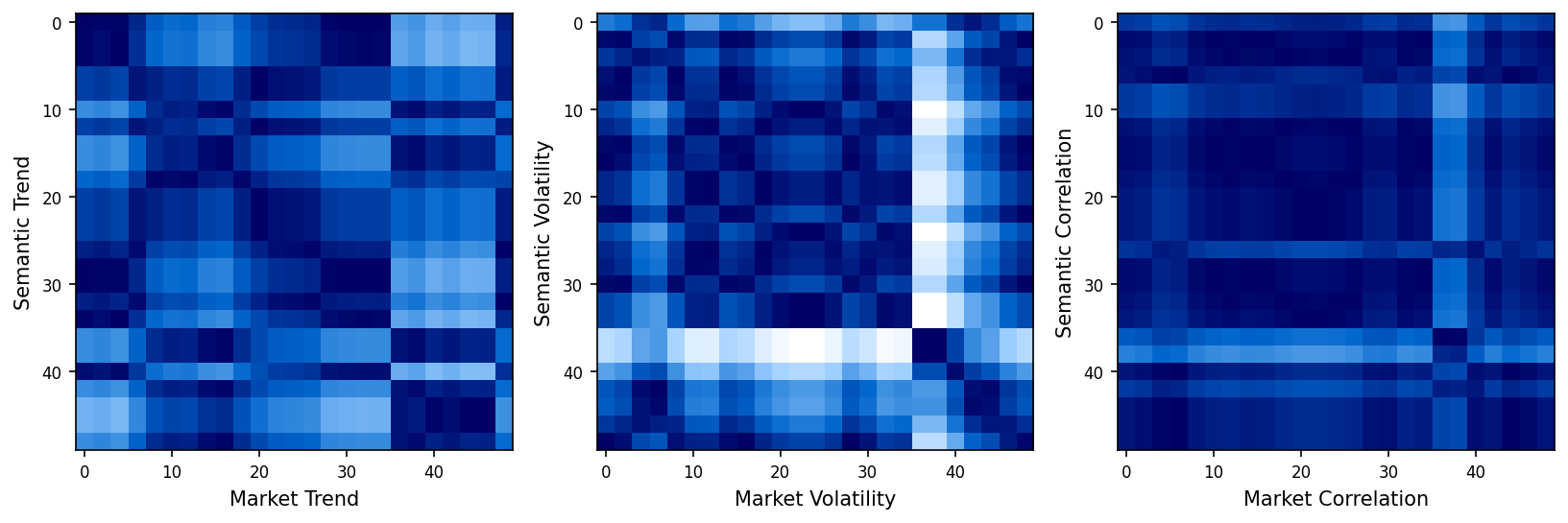}
    \caption{Regime coherence heatmaps for CSI 300 market.}
    \label{fig:regime_heatmaps_csi300}
\end{figure}

\begin{figure}[htbp]
    \centering
    \includegraphics[width=\textwidth]{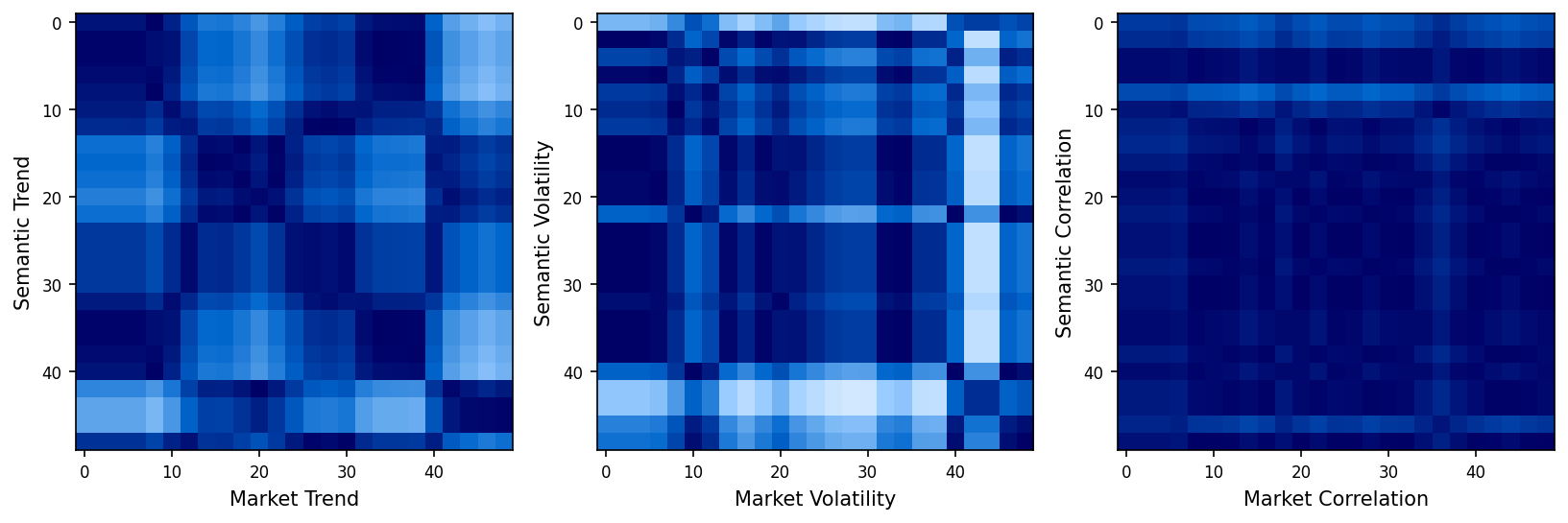}
    \caption{Regime coherence heatmaps for S\&P~500 market.}
    \label{fig:regime_heatmaps_us}
\end{figure}

Figure~\ref{fig:regime_heatmaps_csi300} and Figure~\ref{fig:regime_heatmaps_us} present the resulting heatmaps for the CSI 300 and S\&P~500 markets, respectively. Along the diagonal where semantic assessments align with contemporaneous market proxies, we observe consistently high similarity values (deep blue cells). This indicates that Screener's regime diagnosis, enabled by the tool-based information retrieval and structured LLM assessment, accurately reflects prevailing market conditions for trend and volatility dimensions across most cycles. The off-diagonal regions exhibit lower similarity, confirming that the agent's assessments are primarily responsive to current rather than future or past market states.

The correlation dimension exhibits markedly different behavior. Unlike trend and volatility, which show clear diagonal structures, the correlation heatmaps appear uniformly deep blue across nearly all cycle pairs. This pattern arises because cross-sectional correlations among index constituents remained remarkably stable throughout the evaluation window, with limited temporal variation. Consequently, the market correlation proxy takes nearly constant values across cycles, and Screener's semantic assessments align uniformly with this stable proxy, yielding uniformly high similarity independent of cycle alignment.

In summary, the Screener harness demonstrates reliable regime perception across all three dimensions, with semantic assessments closely tracking contemporaneous market conditions. The strong diagonal alignment observed in the heatmaps confirms that the structured information retrieval and LLM-based diagnosis components of the harness function as intended, grounding qualitative judgments in actual market data.

\subsection{Risk Management and Position Exposure Analysis}
\label{sec:risk_management}

To examine whether AlphaCrafter's integrated harness system dynamically adjusts market exposure in response to prevailing risk conditions, we conduct a case study using a representative trial of Claude Opus 4.6 — selected as the median-performing run to avoid cherry-picking. We focus on the U.S. equity market, where both long and short positions are permissible, in contrast to the long-only constraint of the Chinese A-share market. We construct two rolling-window metrics at a 10-day horizon. The first, \textbf{market volatility}, is defined as the range amplitude over the window:
\begin{equation}
V_t = \frac{\max_{\tau \in [t-9, t]} H_\tau - \min_{\tau \in [t-9, t]} L_\tau}{O_{t-9}},
\end{equation}
where $H_\tau$, $L_\tau$, and $O_\tau$ denote the daily high, low, and open prices of the market index, respectively. The second, \textbf{average net position exposure}, is the 10-day rolling mean of the account's net position rate:
\begin{equation}
E_t = \frac{1}{10} \sum_{\tau = t-9}^{t} r_\tau,
\end{equation}
where $r_\tau$ is the net position rate (long market value minus short market value, divided by total assets) on day $\tau$. Both series are sampled at non-overlapping 10-day intervals to avoid serial dependence, and their relationship is assessed via linear regression.

\begin{figure}[htbp]
\centering
\begin{subfigure}{0.5\textwidth}
    \centering
    \includegraphics[width=\textwidth]{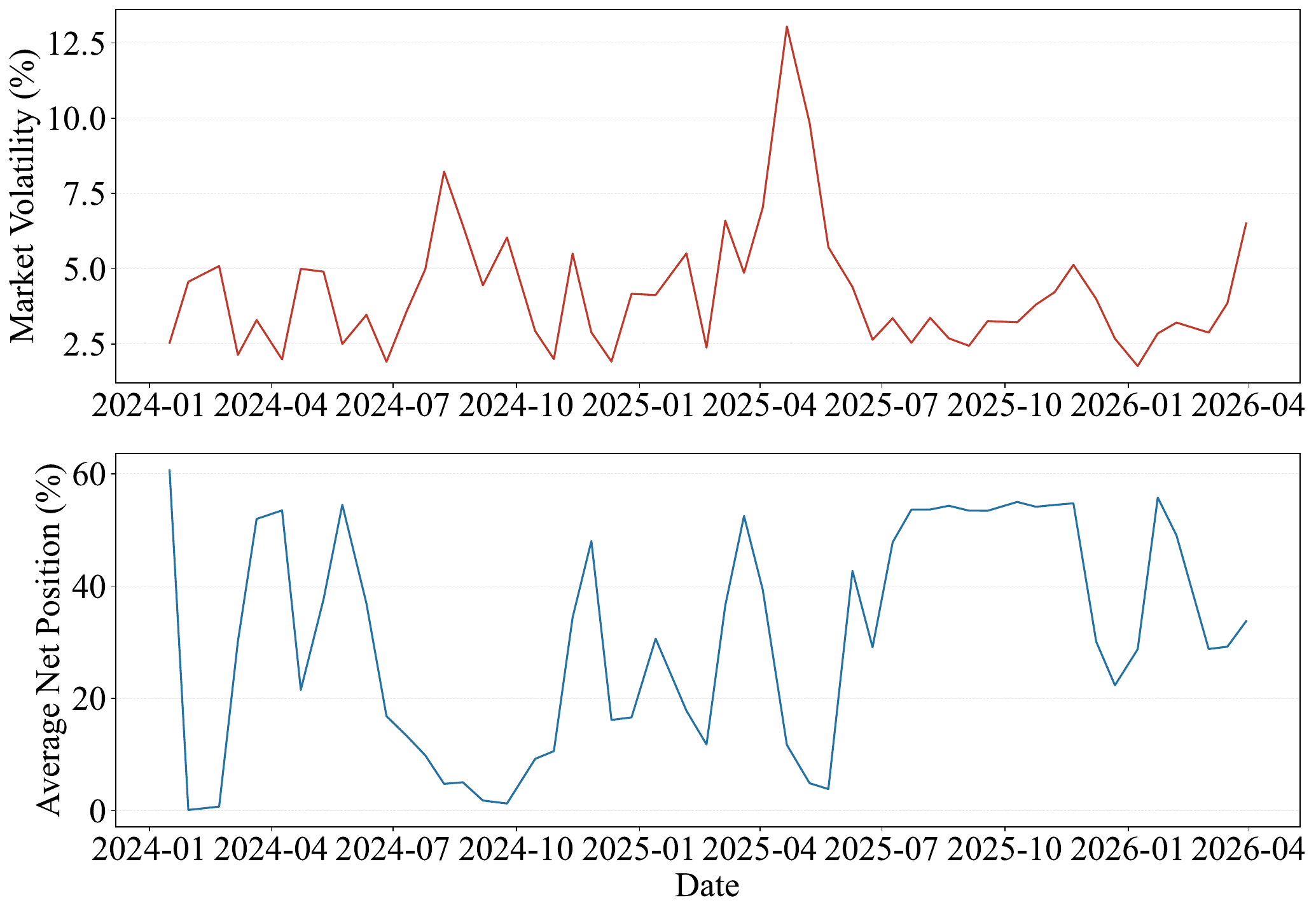}
    \caption{Time series of market volatility and average net position exposure.}
    \label{fig:exposure_vol_ts}
\end{subfigure}
\hfill
\begin{subfigure}{0.4\textwidth}
    \centering
    \includegraphics[width=\textwidth]{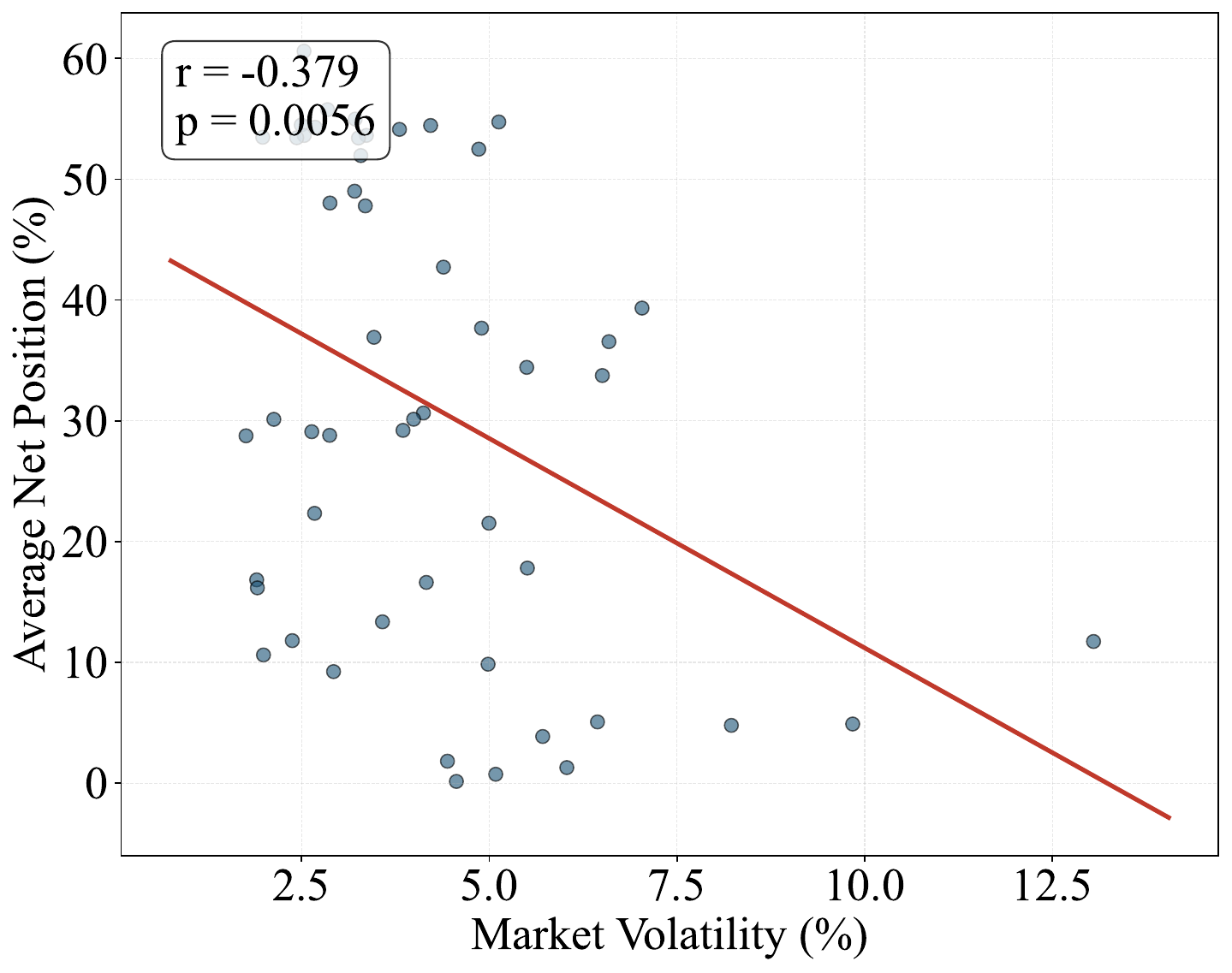}
    \caption{Scatter plot with ordinary least-squares regression line.}
    \label{fig:exposure_vol_scatter}
\end{subfigure}
\caption{Relationship between market volatility and net position exposure for a representative Claude Opus 4.6 trial on the S\&P~500 market.}
\label{fig:exposure_volatility}
\end{figure}

Figure~\ref{fig:exposure_volatility} presents the results. Panel (a) displays the co-evolution of $V_t$ and $E_t$ over the backtest horizon. Visual inspection reveals a clear inverse relationship: during episodes of elevated market volatility, the net position rate tends to decline, whereas calmer regimes coincide with higher exposure levels. Panel (b) corroborates this observation quantitatively. The scatter plot, together with the ordinary least-squares regression line, yields a negative slope, and the Pearson correlation coefficient $r$ confirms a statistically significant inverse association.

This pattern reflects a coherent risk-management behaviour emerging from the AlphaCrafter harness architecture. When market conditions become turbulent, the Screener harness's regime-aware filtering mechanism deprioritizes high-uncertainty factors, while the Trader harness's structured hyperparameter search and backtest validation lead it to reduce net exposure — either by decreasing gross positions or by balancing long and short legs. Conversely, during low-volatility regimes, the system deploys capital more aggressively. Importantly, this adaptive de-risking emerges without explicit volatility-targeting rules encoded in the policies; rather, it is a consequence of the structured interactions among the Miner, Screener, and Trader harnesses, each enforcing its respective constraints and verification mechanisms.

These findings provide evidence that AlphaCrafter's harness-driven design confers a degree of intrinsic risk awareness. The position-taking behavior is not static but responds in a disciplined, counter-cyclical manner to shifting market stress, emerging from the coordinated execution of procedural policies rather than from unconstrained LLM reasoning. This property is particularly valuable in long--short equity markets, where unmanaged gross exposure can amplify drawdowns during volatility spikes. The observed negative correlation between net exposure and market turmoil thus supports the claim that the full harness pipeline yields not only predictive alpha signals but also prudent downside control through its structured, auditable decision processes.

\section{Limitations}
\label{sec:limitations}

While AlphaCrafter demonstrates robust and state-of-the-art performance across multiple markets, it is crucial to contextualize its achievements within the inherent limitations of its design and the experimental setup. Furthermore, these limitations illuminate several promising avenues for future research. This section provides a critical analysis of the framework's current constraints and explores the potential for its evolution into more comprehensive and realistic trading systems.

\textbf{Daily Trading Frequency and Transaction Cost Abstraction.} Our experimental framework is confined to a daily trading frequency, and backtesting is conducted within a simulation environment with abstracted transaction costs. While we mitigate this concern through a dedicated live-trading phase using a real brokerage paper-trading API, the backtesting component necessarily operates under simplifying assumptions that cannot fully replicate the complexities of live markets. Factors such as the price impact of trading illiquid index constituents, time-varying bid-ask spreads, intraday order book dynamics, and episodic liquidity droughts may introduce non-trivial frictions in real-world execution. The abstracted transaction cost model further assumes fixed commission rates and symmetric, zero-mean slippage, which may not hold during periods of market stress or for larger order sizes. Consequently, although our live-trading results provide evidence for real-market viability, the backtesting performance under more realistic market microstructure models remains an open question.

\textbf{Incomplete Mitigation of LLM Data Leakage and Look-Ahead Biases in Backtesting.}
The backtesting phase evaluates AlphaCrafter over historical periods that may partially overlap with the training data of the backbone LLMs. While these models are not explicitly trained on financial time-series for trading purposes, they may have encountered market data, financial reports, or macroeconomic narratives from the backtest window during their large-scale pre-training. This creates a potential data leakage concern: the LLM may possess parametric knowledge of market outcomes during the backtest period, even though such information is not observable to the agent in a forward-looking setting. To address this limitation, we conduct a dedicated live-trading phase using a paper-trading API from a real brokerage, with an evaluation window that falls strictly outside the training data cutoff of all backbone models. This design ensures that live-trading results are free from look-ahead biases. However, the backtesting results should be interpreted with appropriate caution, as they may not fully disentangle genuine reasoning capabilities from memorized market patterns \cite{kong2026evaluatingllmsfinancerequires}.

\textbf{Evaluated Backbone Models and Generalizability.} Our stability study validates AlphaCrafter's performance across three state-of-the-art LLMs: GPT 5.3 Codex \cite{openai2026gpt53codex}, Claude Opus 4.6 \cite{anthropic2026claudeopus46}, and Gemini 3.1 Pro \cite{googledeepmind2026gemini31pro}. While results indicate low sensitivity to the underlying model, this conclusion is drawn from a set of architectures that may share convergent training methodologies, datasets, and alignment techniques. The generalizability of our findings to smaller, open-source models (e.g., Llama series \cite{meta2026llama}, DeepSeek series\cite{deepseek2024deepseek}) or models with distinct pre-training distributions remains unverified. A significant performance cliff when transitioning to more accessible or specialized models would limit the framework's democratization and reproducibility.

\textbf{Scope of the Asset Universe and Extension to Alternative Markets.} The experimental universe encompasses the CSI 300 and S\&P 500 indices over a long period, a setting that already provides meaningful coverage of distinct market structures and macro-financial conditions. Extending this scope to additional asset classes would further enrich the assessment of the framework's generality. In particular, factor definitions, regime dynamics, and execution constraints differ fundamentally in futures, options, and cryptocurrency markets. Exploring the framework's feasibility and adaptation requirements in these alternative markets, especially under the high-leverage and contract-specific microstructure that characterize derivatives trading, constitutes a natural and valuable direction for future work.

\section{Impact Statement}
\label{sec:impact_statement}

This paper presents \textsc{AlphaCrafter}, a multi-agent framework grounded in harness engineering principles that formalizes quantitative trading workflows through structured algorithmic policies. We recognize that automated trading systems carry dual-use potential and warrant careful consideration of their societal implications.

\textbf{Positive Impacts.} By replacing opaque natural language workflows with rigorous algorithmic policy specifications, AlphaCrafter promotes transparency and reproducibility in automated trading system design. The framework lowers the technical barrier to systematic investing by providing a disciplined, modular architecture that smaller research teams and academic institutions can adopt without the extensive infrastructure typically reserved for large financial firms. The regime-aware screening harness enables adaptive portfolio construction that may reduce pro-cyclical herding behavior, potentially contributing to more stable market dynamics during regime transitions. Furthermore, the explicit execution semantics across the entire factor-to-execution lifecycle facilitate regulatory auditing and risk management.

\textbf{Negative Impacts.} Autonomous trading agents, if widely deployed without adequate safeguards, could contribute to market instability through correlated strategy crowding or unintended feedback loops during stress events. The framework's reliance on commercial LLMs raises concerns about the concentration of financial decision-making power in a small number of technology providers. Additionally, while our harness-based design reduces behavioral noise compared to natural language agents, the underlying LLMs may still harbor unexamined biases that propagate through the factor generation and selection process. We encourage future deployments to incorporate circuit breakers, position limits, and mandatory human oversight for live trading, and we advocate for continued research into interpretability and robustness verification for harness-based trading systems.

\end{document}